\pdfoutput=1  
\documentclass[twocolumn,10pt]{article}

\usepackage[margin=0.9in]{geometry}
\usepackage{graphicx}
\graphicspath{{figures/}}
\usepackage{amsmath}
\usepackage{amssymb}
\usepackage{natbib}
\usepackage{float}
\usepackage{xcolor}
\usepackage{booktabs}
\usepackage{dblfloatfix}  
\usepackage[colorlinks=true,linkcolor=blue!50!black,citecolor=blue!50!black,
            urlcolor=blue!50!black]{hyperref}
\usepackage{orcidlink}  
\hypersetup{pdftitle={A learner-free swarm approximates the explore--exploit equilibrium allocation},
            pdfauthor={Galen J. Wilkerson}}

\newif\ifarxivbuild\arxivbuildtrue
\newif\ifsidoc\sidocfalse
\newcommand{\appref}[1]{\ifarxivbuild App.~\ref{#1}\else\ifsidoc Sec.~\ref{#1}\else SI~Sec.~\ref{#1}\fi\fi}

\newcommand{\PaperBibliography}{\bibliographystyle{abbrvnat}\bibliography{refs}}

\begin{document}

\twocolumn[%
  \begin{center}
    {\LARGE\bfseries A learner-free swarm approximates the explore--exploit equilibrium allocation\par}
    \vspace{5pt}
    {\large A self-organized attractor in place of reinforcement-learning machinery\par}
    \vspace{12pt}
    {\large Galen J. Wilkerson\,\orcidlink{0000-0002-7957-5821}\par}
    \vspace{3pt}
    {Komplexity AI LLC \quad \texttt{gwilkerson@komplexai.io}\par}
  \end{center}
  \vspace{8pt}
  \begin{quotation}\noindent
    {\small\textbf{Abstract.} Reinforcement learning (RL) solves the explore--exploit problem by training a learner on reward; its multi-agent,
resource-sharing form has an \emph{equilibrium allocation} as its optimum---the ideal-free distribution (IFD), the
Nash equilibrium of a congestion game. We show a minimal swarm approximates it with no learner at all. We present a
decentralized, learner-free \emph{method} for this problem: agents explore by heavy-tailed (L\'evy) flights,
recruit neighbours to a discovery through a degree-normalized threshold cascade, and disperse as a patch crowds.
The method drives the collective to the \emph{shape} of the equilibrium allocation (pooled spatial matching slope $1.14$, $R^2=0.61$) with
no training, no stored parameters, and no reward in its dynamics; it reaches this as a self-organized attractor
and holds it, re-finding it when the resources move---though it reaches the shape, not the exact equilibrium, and
stays measurably exploitable (Nash-gap $1.46$). Coupling recruitment to movement causes the allocation: removing
it collapses the matching (slope $0.006$). Against Proximal Policy Optimization (PPO) learners
(feedforward, centralized-critic, graph message-passing, and recurrent) trained in the \emph{same} sensing-limited
environment, the method at zero training is the only arm left unbiased w.r.t.\ the IFD gradient (slope
${\approx}1$, CI containing $1$, vs learners $0.69$--$0.97$, all under-concentrating) and ties the best learner's harvest where resource is clustered, while the
learner collects more where resource is scarce ($0.33$ vs $0.22$) and holds a tighter per-capita intake.
Learning buys precision and transfer across reward structures, not a better allocation \emph{shape}. A companion paper
develops the mechanism.
}
  \end{quotation}
  \vspace{14pt}
]

\section{Introduction}

\begin{figure*}[tp]\centering
  \includegraphics[width=0.75\linewidth]{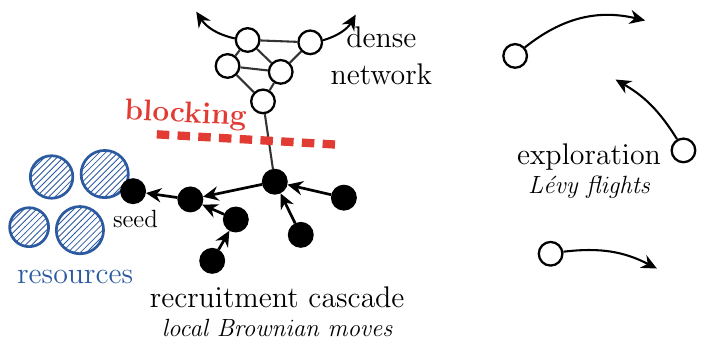}
  \caption{\textbf{The method.} Agents explore by L\'evy flights (right). An agent that lands on a
  resource---a \emph{seed}---triggers a recruitment cascade along the proximity network, drawing neighbours
  into local convergence on the find; as the neighbourhood crowds, degree-normalized activation $a/k$ falls
  below threshold $\phi$ and further recruitment is \emph{blocked} (red). Schematic of the model described in
  \S\ref{sec:mech}; the full model is developed in the companion paper, from which this schematic is adapted~\citep{partI}}
  \label{fig:mech}
\end{figure*}

Drones surveying a region, robot swarms covering an area, software agents dividing work across servers: many
simple units must spread themselves over resources of unknown value, from local information and with no central
controller. When units that converge on the same resource must share it---so that crowding
onto one degrades what each gets---the right spread is a collective \emph{allocation}, and finding it is a
multi-agent explore--exploit problem. Its optimum is not a single best
resource but an \emph{equilibrium allocation}: the ideal-free distribution (IFD), the Nash equilibrium of a
congestion game, at which each unit sits at its best available, crowding-limited
payoff~\citep{fretwell1969ideal,rosenthal1973congestion}. Reinforcement learning (RL) is the machine-learning
account of explore--exploit~\citep{SuttonBarto2018}, and its multi-agent, resource-sharing instance is exactly
this allocation problem. We ask how little a system needs to reach that equilibrium, using a foraging swarm as
the concrete testbed.

\paragraph{A method.} We present a decentralized, learner-free method for this problem. Agents explore by
heavy-tailed (L\'evy) flights; an agent that finds resource recruits its neighbours through a degree-normalized
threshold cascade ($a/k\ge\phi$) on the proximity graph (a random geometric graph, or
RGG~\citep{Penrose2003,DallChristensen2002}); and because activation is degree-normalized, the same
rule disperses the crowd once a patch fills. The method has three fixed controls ($\rho_r,\phi,\alpha$),
self-tunes into its working regime across a $33\times$ density range with no per-problem tuning, and drives the
collective to the \emph{shape} of the equilibrium allocation---foragers in proportion to resource richness
(matching slope $1.14$)---which it reaches as a self-organized attractor (the ideal-free allocation is a
linearly stable fixed point of the rule's coarse dynamics; \appref{app:dynamics}) and holds, re-finding it
when the resources move. Coupling recruitment to movement is what produces the allocation: remove it and
matching collapses (slope $0.006$).

\paragraph{Learner-free and reward-free.} The method does all this with none of reinforcement learning's
machinery---no value function, no policy parameters, no training, and no reward signal anywhere in the loop. No step reads a payoff: a seed fires
on \emph{contact} with resource and recruitment responds to local \emph{crowding}, never to how much an agent
collects. The reinforcement learning is at the level of the \emph{problem solved} and the \emph{allocation
reached}, not the micro-rule---and the tempting reading, that the recruitment rule is a reward-following update
in disguise, is false, which we show directly (\S\ref{sec:how}).

\paragraph{Contributions and claims.} Three contributions: (1) the method---a learner-free, reward-free
coordination rule that drives a multi-agent collective to the optimal allocation's shape
(\S\ref{sec:mech},~\S\ref{sec:solve}); (2) the characterisation that the swarm \emph{reaches} the explore--exploit
equilibrium rather than \emph{doing} RL to get there (\S\ref{sec:how}); and (3) a matched, head-to-head
comparison against trained multi-agent RL under identical embodiment (\S\ref{sec:compare}). Four claims, each
checkable against evidence:
\begin{enumerate}\setlength\itemsep{2pt}
  \item \textbf{It reaches the explore--exploit equilibrium's allocation shape}---the ideal free distribution
        (pooled spatial matching slope $1.14$, a finite-size estimate that relaxes to ${\approx}1$ in the embodied and large-$N$ measurements), caused by the recruitment--movement coupling (remove it and matching collapses
        to $0.006$); the exact equilibrium is not reached, as the swarm stays measurably exploitable
        (\S\ref{sec:solve}).
  \item \textbf{Learner-free and reward-free}---no value function, parameters, or training, and no reward in the
        loop: recruitment is driven by crowding, not payoff (\S\ref{sec:how}).
  \item \textbf{It reaches the allocation, it does not learn}---the configuration's informativeness rises once
        and plateaus, and re-tracks toward the same (never a higher) level after the resources move, with no
        cumulative gain (\S\ref{sec:how}).
  \item \textbf{Benchmarked against trained PPO/MAPPO/GNN/GRU under matched embodiment}---the reward-free swarm is
        the most gradient-faithful arm (unbiased allocation, slope ${\approx}1$ vs $0.69$--$0.97$) and ties the
        best learner's harvest where resource is clustered, but is out-harvested where it is scarce ($0.22$ vs
        $0.33$) and is less precise per capita: learning buys precision and transfer, not a better allocation
        shape (\S\ref{sec:compare}).
\end{enumerate}

\paragraph{Scope of the claim.} We claim the method reaches the IFD in \emph{shape} (foragers in proportion to resource richness at a matching
slope ${\approx}1$, i.e.\ the allocation's \emph{form} rather than the exact equilibrium point), not that its micro-dynamics
are an RL algorithm, a reward-following update, or the replicator equation (the rule is reward-free by
construction; \S\ref{sec:mech}); not that the optimum is a total-reward maximum (under sharing it is an
equal-intake equilibrium; \appref{app:optimum}); and not that the swarm is communication-free---recruitment is
a hard-wired, non-local coordination channel, so the embodied contest isolates the value of a \emph{learner}
given that channel, not the value of the channel itself. The method is for \emph{congestion-structured}
allocation---shared resources whose value degrades with crowding; it reaches the equilibrium in shape, trading
per-capita precision for zero training, and it does not transfer across reward structures. (``Explore''
throughout is \emph{spatial search}---heavy-tailed movement to find \emph{where} resource is---not the epistemic
arm-sampling that drives bandit regret.)

\paragraph{The companion paper.} A companion paper~\citep{partI} gives the mechanism's full
characterisation---phase diagrams over connectivity and threshold, the complete ``cascade window'' map, the
branching analysis, and the harvesting-efficiency results. The present paper is self-contained: the model is
specified in full (\appref{app:modelspec}); the branching ratio is defined and measured in-paper
($\sigma\approx0.067$--$0.086$, subcritical across the $33\times$ range; \S\ref{sec:solve},
Fig.~\ref{fig:selforg}); the slope-one allocation and its ablations are in \S\ref{sec:solve} and
\appref{app:dynamics}; the mean-field stability of the fixed point is derived in \appref{app:dynamics}; and
the movement-matched learner comparison is in \appref{app:embodied}. The companion corroborates; it does not
supply the evidence. What is new in the present paper, beyond the companion's mechanism and phase-diagram
characterisation, is the ideal-free result (\S\ref{sec:solve}), its mean-field stability (\appref{app:dynamics}),
and---the primary contribution---the matched head-to-head against trained multi-agent RL under identical
embodiment (\S\ref{sec:compare}).

\section{Related work}
\label{sec:related}
\paragraph{Explore--exploit and bandits.} The multi-armed bandit~\citep{robbins1952bandit,lattimore2020bandit}
is the canonical formalisation of explore--exploit and the one-state reduction of the general RL
problem~\citep{SuttonBarto2018}. Our setting is its multi-agent, resource-sharing instance: arms are patches
whose per-capita yield falls as more foragers work them, so the optimum is an allocation, not a single best arm.

\paragraph{The optimum: ideal free distribution and habitat games.} The optimal allocation of foragers over
patches is the ideal free distribution (IFD): foragers distribute in proportion to resource so that per-capita
intake is equal and no forager gains by moving~\citep{fretwell1969ideal}. This is the Nash equilibrium of the
habitat-selection (congestion) game~\citep{rosenthal1973congestion,wardrop1952road}---the same allocation that
appears as the Wardrop (user) equilibrium in congested traffic routing and as the proportional-fair (Kelly)
allocation in shared-resource networking~\citep{kelly1998rate}---and the standard optimality
criterion in behavioural ecology. We use the IFD only to \emph{characterise the solution} the swarm reaches, not
to claim it reaches it by playing the game. Empirically, foraging animals typically \emph{under}-match the
resource distribution (matching slope $<1$: richer patches are underused)~\citep{kennedy1993ifd}, which makes
our swarm's slight \emph{over}-matching (slope $1.14$) a departure from the usual ecological bias rather than a
reproduction of it.

\paragraph{Evolved agents reaching the IFD.} The closest precedent is \citet{griffith2006ifd}, where agents
controlled by \emph{evolved} neural networks settle into a near-optimal IFD with no rule explicitly encoding
matching---evidence the IFD can emerge rather than be designed in. It differs from ours in two ways: their
agents carry a learner (a generational evolutionary search tunes the weights), whereas ours has no learner on
any timescale; and they do not compare against a trained reinforcement learner, so what learning adds is left
unmeasured. Theirs is the learner-bearing analogue of ours; ours is its learning-free limit.

\paragraph{Learning in games as the mean field of RL.} Reward-proportional reinforcement is, in the
large-population mean field, the replicator equation~\citep{borgers1997learning}, the bridge used to analyse
multi-agent learning dynamically~\citep{hofbauer1998evolutionary,sandholm2010population,tuyls2005evolutionary,
bloembergen2015evolutionary}. This is the level at which one \emph{might} identify a physical swarm with RL. We
deliberately do not rest our claim there: \S\ref{sec:how} shows the micro-dynamics are reward-free and are not a
replicator update. Our identification is at the problem/solution level. This population-dynamics view also
underlies mean-field multi-agent RL and mean-field games, which make many-agent interaction tractable by
replacing it with each agent's response to the population's average effect~\citep{yang2018meanfield,lasry2007mean};
our coarse mean-field ODE (\appref{app:dynamics}) is a reduction in the same spirit, but it describes the
relaxation of a fixed, reward-free rule toward the ideal-free fixed point rather than the reward-driven learning
updates of many interacting optimizers.

\paragraph{Deep and cooperative multi-agent RL (our trained baselines).} Trained deep-RL agents can \emph{learn}
to forage near-optimally: \citet{wispinski2023foraging} show policy-gradient agents approaching the
marginal-value-theorem patch-leaving optimum---a close precedent for bridging RL and foraging, but there a
\emph{learner} approaches a \emph{single-agent} optimum by training on reward, whereas here a learner-free,
reward-free swarm reaches a \emph{multi-agent} allocation optimum with no training. Our learner is drawn from
modern cooperative multi-agent RL: independent PPO (IPPO)~\citep{dewitt2020ippo}, its centralised-critic
counterpart MAPPO~\citep{yu2022mappo}, and carefully implemented PPO is hard to outperform on cooperative tasks
including foraging~\citep{papoudakis2021benchmark}. We train these as the reward-driven reference against which
the reward-free swarm's precision is measured (\S\ref{sec:compare}). We do not include value-decomposition
baselines (VDN~\citep{sunehag2018vdn}, QMIX~\citep{rashid2018qmix}); these are value-based cooperative methods,
whereas we use the policy-gradient PPO family---a strong and standard baseline on cooperative tasks including
foraging~\citep{papoudakis2021benchmark}---varied across the architectural axes of interest (independent,
centralized-critic, graph, recurrent), which we take as the reward-driven reference.

\paragraph{Non-learning and designed approaches.} Several lines reach optima without a trained reward-maximiser,
by routes different from ours. Metaheuristics---particle-swarm~\citep{kennedy1995particle} and
ant-colony~\citep{dorigo1996ant} optimisation---coordinate simple units but are \emph{designed} to optimise a
supplied objective and carry memory (a personal/global best; a pheromone field read as an external value
estimate~\citep{vellinger2025pheromones}); decentralised control reaches Nash/potential-game equilibria from
local information \emph{by design}~\citep{marden2009cooperative,quijano2010honeybee}; multi-robot coverage control
places agents in proportion to a \emph{known} importance density by descending a locational objective over Voronoi
cells~\citep{cortes2004coverage}---a density-proportional spread like ours, but from an explicit objective and a
sensed field value our agents never read; the ``zero-intelligence''
result shows non-optimising agents reaching a market's competitive allocation---the interaction structure, not
the agent, does the optimising~\citep{gode1993allocative}; and active inference reaches near-optimal behaviour
with no reward by placing an objective \emph{inside} each agent~\citep{friston2009active}. Ours sits at the
reward-free limit: like them it is a designed local rule, but it reads no payoff anywhere in its loop, carries no
learned memory, value estimate, regret signal, or generative model, and its constants are fixed a priori and
demonstrably not harvest-tuned (\S\ref{sec:solve}). No agent minimises the objective; the equilibrium is a
collective property the self-organized dynamics settle at. (We make no entropy-regularised ``soft'' RL
claim~\citep{haarnoja2018soft,ziebart2008maxent}: the converged spread is the IFD set by resource sharing, not a
temperature.)

\paragraph{Decentralized task allocation: thresholds and markets.} Two multi-agent traditions reach an
allocation with no learner, and our rule sits between them. Response-threshold models of insect division of
labour assign a worker to a task when a task-associated stimulus exceeds that worker's threshold, producing
adaptive task fractions with no central controller and no learning~\citep{bonabeau1996quantitative,theraulaz1998response}---the
behavioural account of how harvester-ant colonies regulate foraging from local interaction
rates~\citep{gordon1996regulation,gordon2010ant}. Our recruitment rule $a/k\ge\phi$ is exactly such a threshold,
here \emph{degree-normalized}; what is new is that the ``task'' is a location in a continuous resource gradient
rather than one of a few discrete labour categories, so the thresholds allocate foragers \emph{spatially} and
drive the collective to the ideal-free distribution, not to fixed task proportions. Market- and auction-based
multi-robot task allocation instead coordinates through explicit bids and prices over enumerated
tasks~\citep{gerkey2004taxonomy,dias2006market}; our mechanism has no auctioneer, no bids, no prices, and no task
list---the allocation emerges from a reward-free recruitment contagion on the proximity graph.

\paragraph{Swarm robotics (ground and aerial), and the cost of learning.} Minimal robot swarms self-organize
collective behaviour from local interaction, with no central controller and no learner~\citep{brambilla2013swarm}.
On the ground, identical Kilobots~\citep{rubenstein2012kilobot} assemble into programmed global shapes from purely
local rules~\citep{rubenstein2014selfassembly}, grow reaction--diffusion (Turing) morphologies by neighbour
messaging~\citep{slavkov2018morphogenesis,imrie2017kilobot}, search by random and L\'evy
walks~\citep{dimidov2016randomwalks}, and reach collective decisions or recruit to resources through positive
feedback~\citep{valentini2017bestofn,pitonakova2014recruitment}. In the air, decentralized drone swarms flock and
navigate cluttered real environments from local sensing alone~\citep{vasarhelyi2018flocking,zhou2022wild}. This
tradition engineers behaviour \emph{by design} for a target---a shape, a flock, a consensus---with the rule and
often its parameters tuned or \emph{optimized} to the setting (flocking parameters, for instance, are evolved for
the given environment). Ours is also a designed local rule, but it differs on exactly this axis: its few constants
are fixed a priori, demonstrably not harvest-optimal, and rather than being tuned to a setting the swarm
\emph{self-tunes}---self-organizing to its operating point and holding the ideal-free allocation across a
$33\times$ density range with no per-condition adjustment (\S\ref{sec:solve}). It is also measured against a
resource-sharing optimum and against a trained learner, which this tradition does not do. The reward-free
construction is moreover practically pointed: needing no per-deployment training, the swarm can be launched and
act immediately, spending its (often tightly budgeted) energy on the task rather than on learning---a concrete
advantage for aerial platforms, whose flight time and onboard power are fundamental
constraints~\citep{coppola2020mav}.

\section{The physical model: a self-organizing forager}
\label{sec:mech}

\begin{figure}[t]\centering
  \includegraphics[width={\ifarxivbuild\linewidth\else 3.1in\fi}]{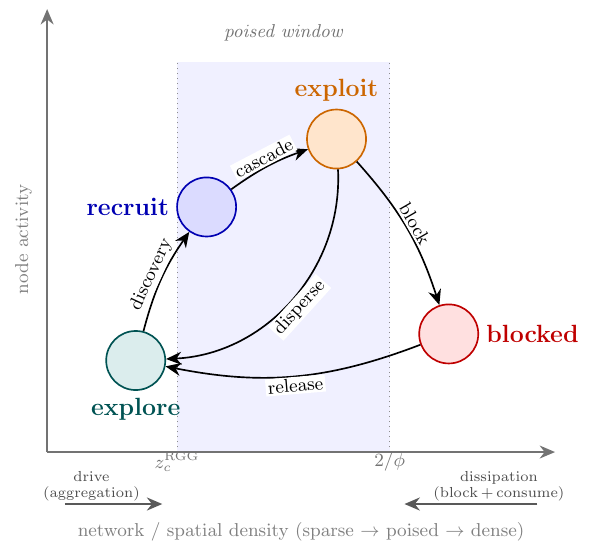}
  \caption{\textbf{The self-organizing cycle balances exploration and exploitation.} In the plane of
  network/spatial density (horizontal) and node activity (vertical), the swarm runs a drive--avalanche cycle:
  discovery seeds a recruitment cascade (explore$\to$recruit$\to$exploit); crowding blocks and disperses it
  (exploit$\to$blocked$\to$explore); aggregation (\emph{drive}) versus dispersal (\emph{dissipation}) hold the
  swarm at a self-organized \emph{quasi}-critical operating point---local clusters near-critical while the
  global branching stays subcritical ($\sigma\approx0.07$)~\citep{bonachela2009soqc,partI}. Schematic of the cycle
  described in \S\ref{sec:mech}; full treatment in the companion paper, from which it is adapted~\citep{partI}}
  \label{fig:cycle}
\end{figure}

$N$ foragers move on a torus. An inactive agent \emph{explores} by heavy-tailed (L\'evy) hops. An agent that
lands on resource becomes \emph{active} (a \emph{seed}) and \emph{recruits} its neighbours: recruitment is a
linear-threshold cascade on the proximity graph (an edge joins agents within sensing range $r$). An agent
activates when the active fraction of its neighbours crosses the threshold, $a/k\ge\phi$ ($a$ active neighbours,
$k$ neighbours), and recruited agents switch to short local moves toward the find. Because activation is
\emph{degree-normalized}, the same rule that spreads a cascade stifles it once a neighbourhood crowds and $a/k$
falls back below $\phi$: recruitment to a crowded patch shuts off, and the swarm releases agents back to
exploration. Crucially, \textbf{no step reads a payoff}: a seed fires on \emph{contact} with resource, and
recruitment responds to the \emph{active fraction} of neighbours---a crowd signal---never to how rich a patch is
or how much intake an agent receives.

Two consequences of the same rule bracket a working range of connectivity. A discovery can recruit beyond a
handful of agents only if the proximity graph \emph{percolates}: below the random-geometric percolation onset
(mean degree $z_c^{\mathrm{RGG}}\approx4.51$) a seed's cascade stays local and dies---a lower edge. At the other
end, because activation needs $a/k$ to clear $\phi$, a too-dense neighbourhood dilutes that fraction and
recruitment stifles, capping the useful mean degree (Part~I locates this blocking edge at ${\approx}2/\phi$,
here ${\approx}11$). Between the edges lies a \emph{cascade window} in which recruitment both spreads and
self-limits; the swarm self-organizes into it and holds there---subcritically, with near-ideal matching---across
a $33\times$ density range with no tuning (\S\ref{sec:solve}, Fig.~\ref{fig:selforg}). The companion
paper~\citep{partI} maps this window in full.

\begin{table}[t]
\centering
{\small
\begin{tabular}{@{}l p{0.38\columnwidth} r@{}}
\hline
\textbf{Symbol} & \textbf{Meaning} & \textbf{Value} \\
\hline
$N$ & foragers & $200$ \\
$L$ & torus side length & $800$ \\
$\rho_r=N(r/L)^2$ & connectivity density & $2.1$ ($r\approx82$) \\
$\phi$ & activation threshold ($a/k\ge\phi$) & $0.18$ ($2/\phi\approx11$) \\
$\alpha$ & L\'evy exponent & $1.0$ \\
$C$ & patches (resource clumps) & $10$ \\
$\beta$ & Pareto patch-richness exponent & $2.2$ \\
$M$ & resource sources & $80$ \\
$z_c^{\mathrm{RGG}}$ & percolation onset & ${\approx}4.51$ \\
\hline
\end{tabular}}
\caption{\textbf{Main model parameters} (spatial foraging model, \S\ref{sec:mech}). The abstract congestion
game (\S\ref{sec:compare}, \appref{app:marl}) uses $K{=}10$ patches and $N{=}200$ agents with the same Pareto
richness; the embodied comparison (\appref{app:embodied}) uses $N{=}200$ over horizon $T{=}600$.}
\label{tab:params}
\end{table}

\paragraph{The method, per step.} Each step, given agent positions and the proximity graph:
\begin{enumerate}\setlength\itemsep{1pt}
  \item \textbf{Sense \& connect.} Each agent senses live sources within $S$; build the proximity graph (edges within $r$).
  \item \textbf{Cascade.} Run the linear-threshold cascade to a fixed point: on-food agents are seeds; an inactive agent with $a/k\ge\phi$ activates; repeat. This labels on-food/recruited agents and their recruitment headings.
  \item \textbf{Move.} Inactive $\to$ L\'evy \emph{explore}; on-food or recruited $\to$ short \emph{approach} step (toward a sensed source, or along the parents' heading).
  \item \textbf{Consume.} Resolve shared consumption (per-capita intake $Q_i/(Nn_i)$), decrement capacities, respawn depleted sources.
\end{enumerate}
No step evaluates a payoff; the configuration carried by positions and the graph---stored nowhere else---\emph{is}
the allocation at that step (full specification, \appref{app:modelspec}).

\paragraph{One rule, two feedbacks.} The method is more than a recruitment rule because two feedbacks run at
once. A \emph{topology--geometry} feedback: cascades move agents, that motion rewires the proximity graph, and
the new graph gates the next cascade~\citep{partI}---a drive--avalanche cycle~\citep{bak1987soc,jensen1998soc}
on a \emph{moving} graph. And a \emph{drive--dissipation} feedback: recruitment is the drive that aggregates
agents out of a too-sparse configuration, degree-normalized blocking the dissipation that disperses them out of
a too-dense one (Fig.~\ref{fig:cycle}). Pushed from both sides, the swarm self-tunes its explore--exploit
balance into a self-organized \emph{quasi}-critical window---local clusters near-critical, the global branching subcritical ($\sigma\approx0.07$)---with no parameter set to put it there---the balance managed in real time by the
physics, not a controller. Recruitment is also \emph{movement-neutral}: recruited agents take short local steps
toward the find (shorter than an exploring L\'evy hop), so no extra movement budget is spent to coordinate.

\section{The swarm reaches the ideal-free allocation}
\label{sec:solve}
\paragraph{The equilibrium.} Coarse-grain the swarm by the fraction $n_i$ of agents committed to patch $i$. With
resource $Q_i$ shared among the $Nn_i$ foragers there, per-capita intake is $f_i=Q_i/(Nn_i)$; the allocation is
stable against relocation when $f_i$ is equal across occupied patches, i.e.\ $n_i\propto Q_i$---the ideal free
distribution, the Nash equilibrium of the congestion game (\appref{app:optimum}). We ask whether the swarm
reaches it.

\begin{figure*}[tp]
  \centering
  \includegraphics[width=\linewidth]{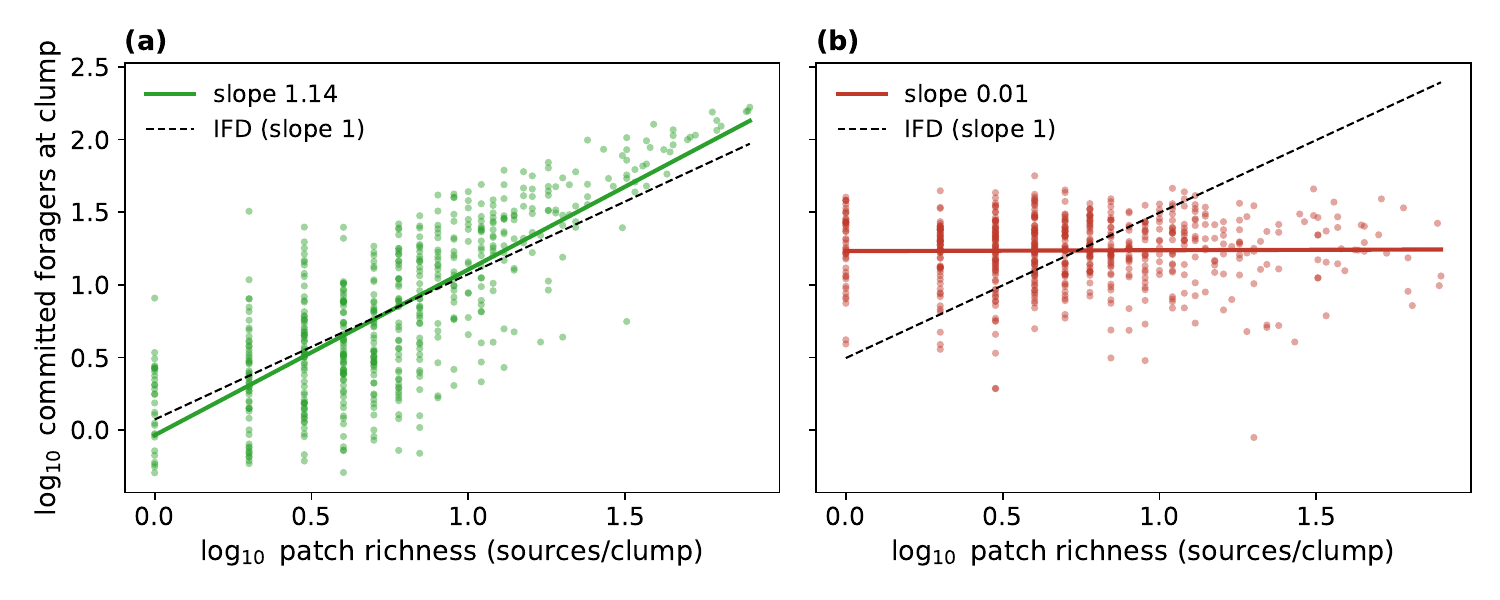}
  \caption{\textbf{The swarm reaches the optimal allocation, and the coupling causes it.} Committed foragers per
  patch vs.\ richness, pooled over $B{=}64$ independent replicas ($n{=}555$ patch-replica points).
  \textbf{(a) Recruiting.} Pooled slope $1.14$ ($95\%$ CI $[1.07,1.21]$ bootstrapping over the $B{=}64$ replicas,
  significantly above the ideal $1$, $p<0.001$; $R^2{=}0.61$)---a systematic \emph{over}-concentration on the
  richest patches, not noise. \textbf{(b) Coupling off.} With recruitment removed, matching is destroyed
  (slope $0.006$): the ideal-free allocation is caused by the recruitment--movement coupling}
  \label{fig:ifd}
\end{figure*}

\paragraph{It reaches the allocation shape, and the coupling causes it.} Pooling committed foragers per patch
against richness over $B{=}64$ replicas, the recruiting swarm matches with log--log slope $1.14$
(Fig.~\ref{fig:ifd}a; $95\%$ CI $[1.07,1.21]$ over the $B{=}64$ replicas, significantly above $1$, $p<0.001$)---a systematic \emph{over}-concentration on the richest patches. Removing the recruitment--movement
coupling destroys the matching (slope $0.006$; Fig.~\ref{fig:ifd}b): the allocation is \emph{caused} by the
coupling, not a property of the individual agents. (``Coupling off'' removes only the recruited agents'
\emph{heading}: they still slow to short steps but move in a random direction, while seeds still perform
area-restricted search on food and inactive agents still explore by L\'evy flight. Discovery is by
contact---the model has no individual gradient-climb toward sensed-but-unreached resource---so removing
recruitment-convergence is precisely what flattens occupancy against richness.) This collapse is specific: the allocation shape is robust to
the swarm's exploration controls (the L\'evy exponent out to its Brownian limit, and the activation threshold;
\appref{app:robust}), so it is the recruitment--movement coupling, not foraging activity as such, that produces
the proportional allocation. A second ablation localizes the mechanism: removing the
crowding cutoff (the degree-normalized blocking) leaves a near-ideal slope that does not collapse
($1.14\to0.95$), though the swarm tracks the ideal less tightly (total-variation distance $0.22\to0.37$) and
commits more of itself to patches (active fraction $0.71\to0.86$; \appref{app:dynamics}). So the two ablations
separate two roles: the allocation \emph{shape} is set by richness-proportional discovery and recruitment (it
survives removing the cutoff but collapses without the coupling), while the crowding cutoff sharpens the match
and regulates \emph{where} on the explore--exploit balance the swarm sits. The allocation is reached
\emph{dynamically}---the coupled swarm climbs from near zero to its plateau by ${\sim}200$--$260$ steps and then
holds, the uncoupled swarm stays flat (\appref{app:dynamics}, Fig.~\ref{fig:convergence}).

\begin{figure*}[tp]\centering
  \includegraphics[width=\linewidth]{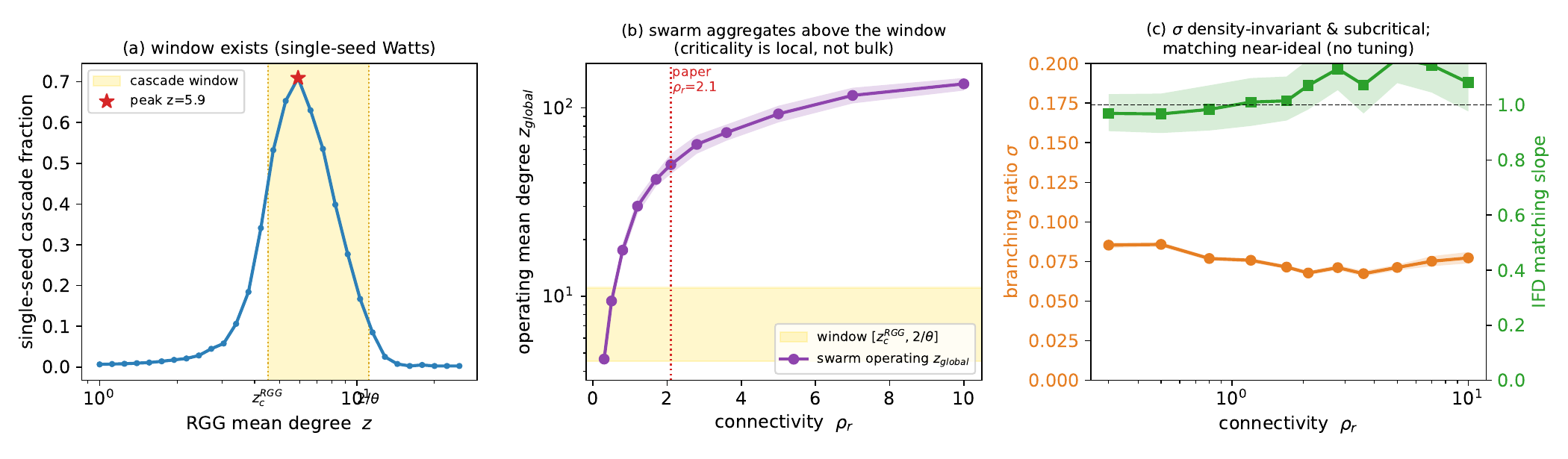}
  \caption{\textbf{The operating regime is self-organized, not tuned.} \textbf{(a)} The single-seed recruitment
  cascade has an inverted-U window between $z_c^{\mathrm{RGG}}\approx4.51$ and the blocking bound $2/\phi$ (peak
  near $z\approx5.9$). \textbf{(b)} Swept over a $33\times$ connectivity range, the mobile swarm's operating
  (bulk) mean degree runs \emph{above} the single-seed window---it aggregates---so the criticality is local, not
  bulk. \textbf{(c)} Across the same range and with no per-condition tuning, the branching ratio stays subcritical
  and nearly flat ($\sigma\approx0.067$--$0.086$, left axis) while the IFD matching stays near-ideal (slope
  $0.97$--$1.17$, right axis)}
  \label{fig:selforg}
\end{figure*}

\paragraph{The operating regime is self-organized, not tuned.} The swarm is not placed at a hand-chosen
operating point. Sweeping connectivity over a $33\times$ density range ($\rho_r$ from $0.3$ to $10$; Fig.~\ref{fig:selforg}), it holds a
\emph{subcritical} branching ratio $\sigma$---the mean number of further agents each freshly-recruited agent
recruits, so $\sigma<1$ means cascades are finite---nearly flat ($0.067$--$0.086$), and maintains near-ideal
matching throughout (slope $0.97$--$1.17$) with no per-condition tuning; a full Sobol sweep confirms the swarm's
own controls $\phi,\alpha$ sit on broad plateaus (\appref{app:robust}). ``Not tuned'' is about fitting to the
task, not being parameter-free: $\alpha$ and $\phi$ set \emph{where} in the regime the swarm operates, are held
at a-priori values, and neither is at its harvest-optimal value---so the swarm meets the learner
(\S\ref{sec:compare}) with parameters never fitted to the task. (The criticality is \emph{local}, not global: individual clusters sit near-critical while the swarm's bulk
degree runs above the single-seed window $[z_c^{\mathrm{RGG}},2/\phi]$ because recruited agents aggregate, so the
global branching stays subcritical---self-organized \emph{quasi}-criticality~\citep{bonachela2009soqc},
characterized in full in Part~I~\citep{partI}.) The matching is also robust to system size: grown over a $16\times$ range of $N$
with every dimensionless control held fixed, the slope converges to ${\approx}0.94$ as $N$ grows while harvest
stays flat (\appref{app:robust}, Fig.~\ref{fig:nscaling})---the $N{=}200$ values used throughout are
representative.

\section{How it reaches the allocation: learner-free, reward-free, and not by learning}
\label{sec:how}
The swarm reaches the allocation; this section establishes \emph{how it does not}---without a learner, without a
reward signal, and without learning over time---showing each directly.

\paragraph{No learner.} There are no adjustable parameters and no training phase. Ordinary
RL~\citep{SuttonBarto2018} keeps a value function or policy and updates it over episodes; the trained parameters
are the product. Here the instantaneous configuration $\mathbf n(t)$, carried by positions and the interaction
graph and stored nowhere else, \emph{is} the allocation, taken in real time, with nothing trained or stored.

\paragraph{No reward signal, by construction.} Reward-freeness is a property of the rule's construction: per-capita intake
$f_i$ is never computed or read---a seed fires on contact, recruitment responds to the active \emph{fraction} of
neighbours (\S\ref{sec:mech}). There is no reward in the loop to follow; $f_i$ is a quantity \emph{we} compute
to describe the outcome. This is why agents pile onto richer patches only because richer patches, having more
sites, are discovered and re-seeded more often---not because anyone measures that they pay better. The
mechanism is thus reward-\emph{free} but not value-\emph{blind}: resource value enters through discovery
frequency, never through a payoff an agent reads. Logged flow
is consistent with this (inflow tracks occupancy, not payoff), but under equal sharing such correlations are
non-identifying, so the proof is the construction, not the measurement (\appref{app:dynamics},
Fig.~\ref{fig:rewardfree}).

\paragraph{A relaxation, not a reward-gradient.} Is the flow nonetheless a reward-following law in disguise? No.
Fitting the denoised mean flow over the transient to competing mean-field laws by cross-validation
(\appref{app:dynamics}), a reward-proportional \emph{replicator} update is not the best description (held-out
$R^2=0.26$); the flow is a nonlinear \emph{relaxation} toward the ideal-free fixed point (held-out $R^2$ up to
$0.64$). The coarse dynamics relax the allocation to the equilibrium; this is a characterisation of the flow,
not an RL identification.

\paragraph{It reaches the allocation, it does not learn.} The swarm does \emph{adapt}: its configuration---agent
positions, the proximity graph, and each agent's mode (exploring, recruited, on food, blocked)---shifts to track
where resource currently is, so it carries information about the field and indeed \emph{is} the allocation
(\appref{app:modelspec}). What it does not do is \emph{learn} in the reinforcement sense of improving with
experience: no value estimate, reward count, policy, or memory accumulates past outcomes to bias future action
toward higher return; the configuration is re-formed each step by the fixed rule from current local information.
Behaviourally (\appref{app:dynamics}, Fig.~\ref{fig:plateau}): under fixed conditions the configuration's
informativeness about the resource field rises once and \emph{plateaus} (by ${\sim}163$ steps---an information-theoretic settling time, distinct from the allocation-slope plateau of ${\sim}200$--$260$ steps in \S\ref{sec:solve}) rather than
improving with experience; and when the resources are relocated mid-run the swarm re-tracks \emph{toward} the
same quality it held before---recovering within the window but never climbing above it---with no cumulative gain.
(The informativeness metric is bounded, so the plateau alone is only suggestive; the load-bearing evidence is the
absence of any reward-driven value or policy plus the re-tracking toward the same, not a higher, level.) The
swarm reaches and \emph{holds} the equilibrium, re-finding it when the environment changes---a converged,
homeostatic policy, not an agent accumulating knowledge.

\section{Comparison to a trained learner: who reaches the allocation}
\label{sec:compare}
How does a learner-free method compare to an actual learner? We compare in two settings: an \emph{abstract}
(disembodied) congestion game---direct patch choice with a payoff oracle and no spatial search, against
independent softmax policy-gradient and $Q$-learning agents (\appref{app:marl})---and the \emph{same embodied
environment} the swarm acts in, where the only difference between arms is fixed versus learned control
(\appref{app:embodied}). The embodied comparison is the decisive one.

\begin{figure*}[tp]\centering
  \includegraphics[width={\ifarxivbuild 0.7\linewidth\else\linewidth\fi}]{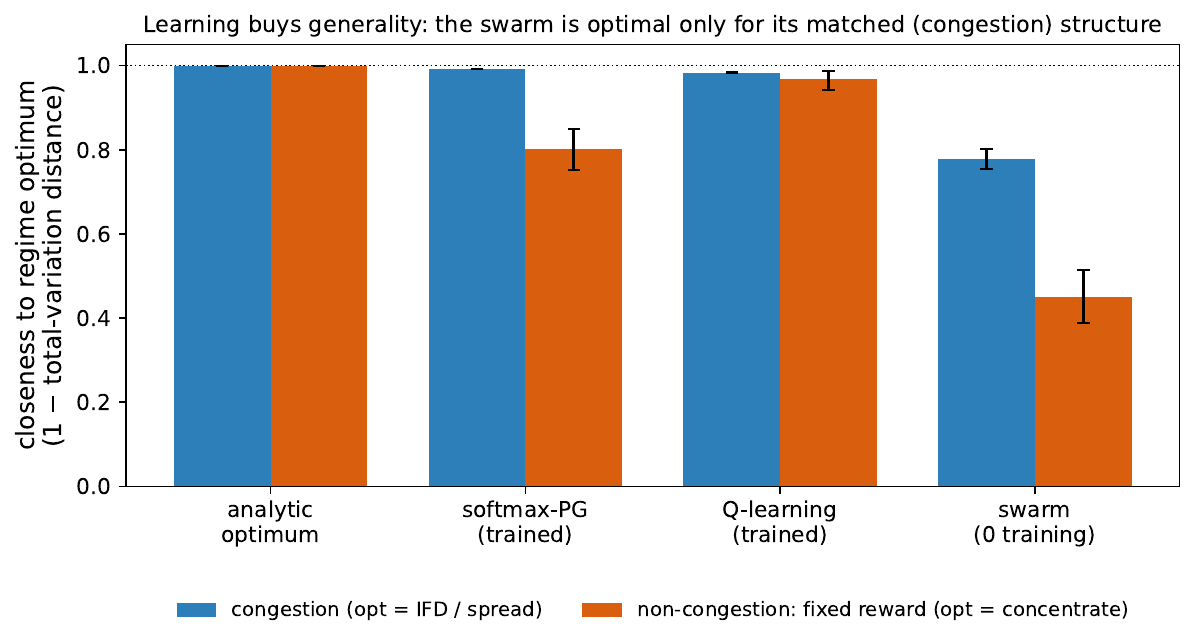}
  \caption{\textbf{Transfer is what a learner buys.} Closeness to the regime optimum in the \emph{congestion}
  regime (optimum $=$ ideal free distribution) and a \emph{non-congestion} regime (fixed per-patch reward,
  optimum $=$ richest patch). The reward-free swarm is near-optimal only in the congestion regime its mechanism
  matches; the learners, reading reward, adapt to both. Abstract congestion game; see \appref{app:marl}}
  \label{fig:transfer}
\end{figure*}

\paragraph{Disembodied: the learner buys precision, and transfer.} Handed a payoff oracle and direct patch
choice, a trained learner reaches an exploitability (per-capita intake CV; $0$ at the equal-intake Nash point) the swarm lacks: the swarm reaches the right allocation
shape at zero training (matching slope $1.275$, consistent with the spatial $1.14$; $78\%$ of the way to Nash)
but its Nash-gap ($1.46$) stays orders of magnitude above a converged learner's (\appref{app:marl},
Fig.~\ref{fig:marlabstract}). That precision is a feature of the \emph{easy} game, and it does not carry to the
embodied setting below. Learning also buys \emph{transfer}: in a non-congestion regime (fixed per-patch reward,
optimum $=$ the richest patch) the learners adapt ($80$--$97\%$ of optimum) while the reward-blind swarm keeps
spreading and is sub-optimal ($45\%$; Fig.~\ref{fig:transfer})---a learner reads reward and transfers across
reward structures; the fixed, reward-blind method cannot. This is the method's honest operating boundary.

\paragraph{Embodied and sensing-limited: most gradient-faithful, on par on harvest only where resource is clustered.}
The decisive comparison matches primitives so neither arm can act on what it cannot sense: the \emph{approach}
action steers an agent only toward resource sensed within radius $S$ (or, if recruited, along its neighbours'
heading), identical for every arm. We score on two fields---\emph{scarce} (sparse, hard-to-find) and
\emph{clustered} (patchy, resource-rich)---by the fraction of a validated oracle ceiling (the physical maximum intake under free placement, not optimal play), and on two allocation
measures: the ideal-free slope (how faithfully foragers track richness) and the exploitability (per-capita
intake CV, $0$ at the equal-intake Nash point). We train three learner families to convergence
($150{,}000$ env-steps, $16$ seeds): feedforward IPPO; a communication-capable graph policy running learned
message passing over the same proximity graph the cascade uses; and, as the steelman, MAPPO (a centralized
critic that sees privileged global state). Each gets a rich $14$-d observation and a dense shaped reward; only
the controller differs from the swarm (\appref{app:embodied}).

\emph{Harvest.} In the clustered field the reward-free swarm at zero training is statistically \emph{equivalent}
to the best trained learner ($0.80$ vs MLP-IPPO's $0.80$; a two-one-sided-test rejects a difference beyond
$\pm0.02$, $90\%$ CI $[-0.004,+0.008]$, $p<0.001$), and is not out-harvested by the MAPPO steelman ($0.76$). The $\pm0.02$ equivalence margin is roughly $3\%$ of
the random-baseline-to-ceiling harvest span (random $0.32$ to the oracle ceiling $1$)---a substantive scale,
not one chosen to clear the test.
Both arms plateau at the same ${\approx}0.80$---the matched-sensing level the strongest arms reach (MAPPO $0.76$, GNN $0.78$, GRU $0.75$ trail), which a
$4\times$-capacity/learning-rate sweep moves by ${<}0.01$ (\appref{app:embodied}), so the equivalence is not
against an under-powered baseline. In the scarce field the learner collects more ($0.33$ vs $0.22$, $p<0.001$;
random $0.14$): where resource is hard to find, reading reward and training buys raw harvest the reward-free
method does not. We do not claim the swarm out-harvests a learner universally; it ties the best learner where
resource is clustered and trails where it is scarce.

\begin{figure*}[tp]\centering
  \includegraphics[width=\linewidth]{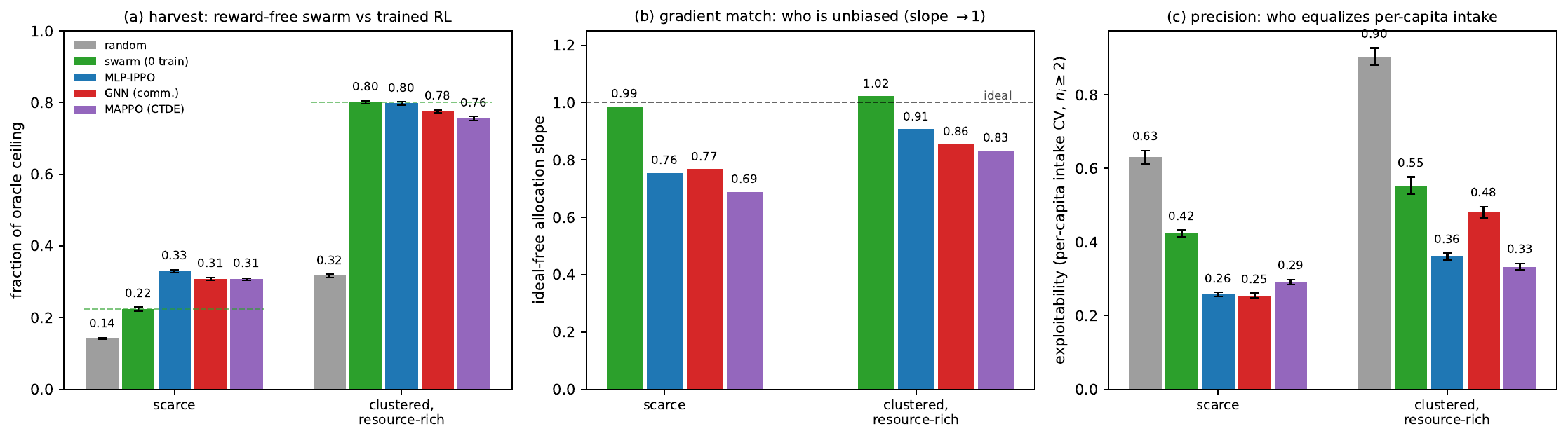}
  \caption{\textbf{Embodied comparison under matched, sensing-limited primitives.} Same environment;
  \emph{approach} steers only toward resource sensed within $S$, identical for every arm; $16$ seeds, trained to
  convergence; the communicating learner is a one-round ($L{=}1$) graph policy ($L{=}3,5$ do not help), MAPPO a
  centralized-critic learner. \textbf{(a)} Harvest (fraction of oracle ceiling): clustered---swarm $0.80$
  (zero training) $\equiv$ MLP-IPPO $0.80$ (TOST, $\pm0.02$) $>$ MAPPO $0.76$, GNN $0.78$, random $0.32$;
  scarce---swarm $0.22 <$ MLP-IPPO $0.33$, random $0.14$. \textbf{(b)} Ideal-free slope (dashed $=1$): the swarm
  is the only arm whose slope CI contains the ideal $1$ (slope $0.99,1.02$); every learner under-concentrates ($0.69$--$0.97$, GRU closest at $0.97$), and the swarm's lead over the IPPO comparator excludes $0$.
  \textbf{(c)} Exploitability (per-capita CV; lower $=$ tighter): the ordering reverses---the learner equalizes
  per-capita intake more tightly, the swarm nonetheless well below random. (b) and (c) are the bias--variance
  split: emergent allocation is gradient-unbiased but noisier}
  \label{fig:stronger}
\end{figure*}

\emph{Allocation: a bias--variance split.} On the \emph{shape} of the allocation---how faithfully foragers track
richness---the reward-free swarm is the only arm whose slope CI contains the ideal $1$ (slope $0.99$ scarce [$95\%$ CI $0.96,1.02$], $1.02$ clustered
[$0.98,1.07$]; seed-cluster bootstrap over $16$ seeds, Fig.~\ref{fig:stronger}b)---it is gradient-unbiased, while
every trained learner under-concentrates (point estimates below $1$: IPPO $0.76$/$0.91$; MAPPO $0.69$/$0.83$;
GNN $0.77$/$0.86$; recurrent GRU $0.97$ clustered, the closest learner; Fig.~\ref{fig:stronger}b). Its lead over the
feedforward IPPO comparator is statistically resolved ($+0.23$ [$0.21,0.27$] scarce, $+0.11$ [$0.06,0.17$]
clustered, both CIs excluding $0$); the swarm's clustered edge over the closest learner (GRU) is only $0.05$ and
within noise, so the load-bearing claim is unbiased-versus-under-concentrating, not a strict slope ordering over
every learner. On the \emph{precision}---how tightly
per-capita intake is equalized---the ranking reverses: the learner is tighter, though the swarm stays well below
random (Fig.~\ref{fig:stronger}c). The two do not contradict: an allocation can sit on the richness gradient on
average yet scatter patch-to-patch, and the cascade's heavy-tailed occupancy produces that scatter. The honest
reading is a bias--variance split (an organizing analogy, not a formal decomposition): \emph{the reward-free
method reaches the correct allocation shape at zero training and with no reward, more faithfully than any
learner tested; learning buys precision, not a better match to the gradient}.

\paragraph{Neither communication nor a centralized critic rescues the learner.} Two objections---that a
feedforward learner cannot represent the cascade's within-step coordination, and that independent learners lack
joint credit assignment---are addressed. Learned message passing does not help (the graph policy reaches $0.78$
clustered, below both the swarm and feedforward IPPO; extra rounds are no better), and the shortfall is not its
smaller environment batch: re-running at a matched count cut its gradient variance to the feedforward learner's
yet left it below the swarm (\appref{app:embodied}). The MAPPO centralized critic, with privileged global
state at training, reaches only $0.76$ clustered. We read these as a bound on these learners at this budget, not
proof no architecture could---but together they remove the obvious ``the learner merely lacked communication or
joint credit assignment'' objections. Nor does \emph{learnable} memory: a recurrent GRU-IPPO learner, trained to
convergence on the same budget, falls in the same pattern (clustered harvest $0.75$, below the swarm;
under-concentrated, slope $0.97$; tighter per-capita precision; \appref{app:embodied}).

\paragraph{The under-concentration is not a shaping artifact.} A natural objection is that the learner
under-concentrates only because its reward shaping and entropy bonus push it to spread. Retraining the binding
comparator with the proximity-shaping term removed and the entropy coefficient annealed to zero, the
under-concentration persists (slope stays $0.76$--$0.78$ scarce, $0.86$--$0.92$ clustered; and removing the
shaping if anything slightly \emph{lowers} the clustered slope, the opposite of the predicted spreading;
\appref{app:embodied}). It is intrinsic to what the learner can represent and optimize under matched sensing,
not an artefact of the shaping.

\paragraph{Across the patchiness axis.} Sweeping the field's patchiness over ${\sim}5$ points turns the
two-field claim into a curve (\appref{app:robust}): at the most clustered field the swarm matches the learner
on harvest with a slightly steeper slope, and the learner's harvest edge opens monotonically as the field is
made diffuse, both declining toward the random floor at the most diffuse. The swarm is competitive---and in
allocation shape slightly ahead---precisely where the field is clustered enough for cascades to form, and cedes
raw harvest as that structure disappears. This is one pair of fields and four learner families: it
\emph{bounds} rather than settles the general swarm-vs-RL comparison.

\begin{table*}[tp]
\centering
{\small
\begin{tabular}{@{}p{0.30\linewidth}p{0.32\linewidth}p{0.30\linewidth}@{}}
\hline
\textbf{To reach the allocation} & \textbf{Learner-free swarm (this work)} & \textbf{Trained multi-agent RL} \\
\hline
Reward signal & none (reward-free) & per-agent scalar payoff, every step \\
Stored parameters & none & a policy (and critic) network \\
Training before use & none; acts at its operating point & $150{,}000$ env-steps, dense shaped reward \\
Parameters tuned to the task & none; $\alpha,\phi$ fixed a priori, not harvest-optimal & policy fit to a dense reward \\
Harvest, clustered field & $0.80$ of ceiling at \emph{zero} training & $0.80$ (IPPO): equivalent (TOST); $0.76$ (MAPPO) \\
Harvest, scarce field & $0.22$ & $0.33$ (IPPO): learner collects more \\
Allocation shape (gradient match) & slope ${\approx}1$ both regimes (most faithful) & slope $0.69$--$0.97$ (under-concentrates) \\
Allocation precision (per-capita CV) & $0.42$/$0.55$: below random, above learner & $0.26$/$0.36$ (IPPO): tighter \\
Transfer across reward structures & none (congestion/IFD only) & adapts \\
\hline
\end{tabular}}
\caption{\textbf{The two pay for different things.} The swarm reaches the allocation shape with no reward,
parameters, or training; the learner spends all three to buy precision and transfer. See \S\ref{sec:compare},
Fig.~\ref{fig:stronger}.}
\label{tab:cost}
\end{table*}

\section{Discussion}
\paragraph{Where the reinforcement learning is.} The identification is at the level of the \emph{problem} and
the \emph{outcome}, not the micro-rule: the swarm reaches the explore--exploit equilibrium's allocation, while
its dynamics read no reward and are not a learning update. A fixed physical mechanism occupies the solution a
learner is trained to find.

\paragraph{Significance.} That a minimal, non-optimising system approximates an ideal-free allocation with no
learner is the headline; the matched comparison sharpens it into a split---under identical embodiment the
reward-free method is the most gradient-faithful arm and matches the best learner's harvest only where resource
is clustered (it is out-harvested where resource is scarce), while learning buys precision and transfer.
As a method, it is a zero-training, reward-free, decentralized solution for congestion-structured allocation.

\paragraph{Reconciling the matching slopes.} The matching slope is reported in several settings; they are
consistent once one accounts for what each measures (Table~\ref{tab:slopes}). All lie near the ideal $1$
($0.94$--$1.28$)---foragers in proportion to richness---and the spread tracks the measurement: the pooled-spatial
estimate over-concentrates slightly ($1.14$) and the disembodied game slightly more ($1.275$), while the
single-field embodied measurement and the large-$N$ limit land at essentially $1$ ($0.99$--$1.02$; $0.94$).

\begin{table*}[t]
\centering
{\small
\begin{tabular}{@{}p{0.34\linewidth} l p{0.40\linewidth}@{}}
\hline
\textbf{Setting / field / measurement} & \textbf{Slope} & \textbf{What it measures / why} \\
\hline
Spatial foraging, pooled over $B{=}64$ replicas (committed foragers vs.\ richness; \S\ref{sec:solve}, Fig.~\ref{fig:ifd})
 & $1.14$ & $95\%$ CI $[1.07,1.21]$, $R^2{=}0.61$; pooled occupancy, mild \emph{over}-concentration on the richest patches \\
Abstract congestion game, disembodied---direct patch choice, payoff oracle, no spatial search (\S\ref{sec:compare}, \appref{app:marl})
 & $1.275$ & consistent with the spatial $1.14$; $78\%$ of the way to Nash \\
Embodied, matched sensing-limited primitives, per field, $16$ seeds (\S\ref{sec:compare}, \appref{app:embodied})
 & $0.99$ / $1.02$ & scarce / clustered; essentially the ideal slope $1$ \\
Cutoff-removed ablation---degree-normalized blocking replaced by an absolute-count rule (\appref{app:dynamics})
 & $0.95$ & allocation shape survives removing the cutoff; the match loosens (TV $0.22{\to}0.37$) \\
Large-$N$ limit, all dimensionless controls fixed, $N{=}50$--$800$ (\appref{app:robust}, Fig.~\ref{fig:nscaling})
 & $0.94$ & the small-$N$ over-concentration damps with size; $N{=}200$ is representative \\
\hline
\end{tabular}}
\caption{\textbf{Matching slope by setting and measurement.} Every setting gives a slope near the ideal $1$
($0.94$--$1.28$): foragers in proportion to resource richness. The spread is a measurement difference, not an
inconsistency---pooling committed foragers across replicas captures the heavy-tailed cascade occupancy that
over-concentrates on the richest patches (a finite-size effect that damps toward $0.94$ as the system grows),
whereas the embodied per-field measurement under matched, sensing-limited primitives reads the slope at
essentially $1$}
\label{tab:slopes}
\end{table*}

\paragraph{Limitations.} (i) The equilibrium reached is the ideal free distribution of a resource-sharing
problem---the common case when resources are shared, but not every allocation problem. (ii) Two forces set the
allocation and we separate them: blocking regulates the explore--exploit \emph{balance} and operating point
(shown here across a $33\times$ density range, mapped in full in~\citep{partI}), while the allocation
\emph{shape} is set by richness-proportional discovery and recruitment---the cutoff-removal ablation
($1.14\to0.95$, \appref{app:dynamics}) makes the separation explicit. (iii) The trained-learner comparison is
one pair of fields and four learner families: it \emph{bounds} rather than settles the general comparison,
though the centralised-critic steelman does not overturn it. (iv) We give no closed-form regret or
convergence-rate bound; the stability result is mean-field, and the reduced form's fixed point coincides with the
ideal free distribution by its choice of exponents ($s=p=1$), so it is a consistency check on the coarse dynamics,
not an independent derivation---nor a theorem for the microscopic stochastic rule---of why the swarm reaches the IFD.
(v) The trained learners span feedforward, graph-message-passing, centralized-critic, and recurrent (GRU)
policies (all also carrying hand-crafted short-memory features); every one under-concentrates relative to the
swarm and none exceeds its clustered-field harvest at this budget---which bounds rather than settles whether some
other architecture or training budget could. The binding comparators are already plateaued (relative gain
${<}0.3\%$ over the final quarter in both fields; \appref{app:embodied}), so more training at these architectures
is unlikely to close the clustered-field gap, though we do not run longer budgets to prove it.

\paragraph{Computational cost.} At deployment all methods are linear in the agent count $N$ per step: the
swarm's work is $\Theta(N)$ (bounded-degree sensing and recruitment; the subcritical cascade propagates
$\Theta(N/(1-\sigma))=\Theta(N)$ work), the same order as a learner's forward pass. The difference is
\emph{offline}: the swarm trains on zero samples and stores no parameters, while each learner consumes
$150{,}000$ env-steps and a stored policy. Cost is not where the swarm loses: it is strictly cheaper offline and
no worse than linear online. What a learner's training budget buys is the scarce-field harvest edge, tighter
per-capita precision, and transfer---not a cheaper deployment.

\paragraph{Scope, outlook, and a target application.} Nothing in the mechanism is specific to food: it needs
only many simple units with local sensing and communication, resources whose value degrades as units crowd them,
and a need to allocate without a central controller---drone and robot swarms over survey regions, software
agents across servers, distributed load balancing. We demonstrate none of these here, and as the non-transfer
result shows the mechanism does not carry to reward structures it does not encode, so whether it transfers (and
with what loss of precision) is an open empirical question per domain. One concrete target that fits this profile is
battery-limited multicopters mapping the urban heat
island~\citep{duffy2021drones,lemonsu2026paname,roberts2025uas}, where sampling value is local thermal
variability that \emph{saturates} as drones crowd a region---an ideal-free allocation problem whose negative
feedback is information saturation rather than depletion; a learner-free fleet could concentrate flight time in
proportion to local variability, online and within battery, with no training or prior map. The theoretical next
steps are to characterise \emph{where} the learner's harvest edge appears and to give a convergence-rate or
regret characterisation of the physical flow. The mechanism also leaves room for other simple, emergent
refinements that could make the swarm more efficient without adding a learner. One is a per-hop signal
decay along the recruitment cascade: here the cascade is undecayed (as in the companion paper~\citep{partI}),
but tuning how recruitment strength attenuates with cascade depth is a natural efficiency knob---one such
refinement among others.

\section{Conclusion}
A foraging swarm faces the explore--exploit problem that defines reinforcement learning. We have shown that a
minimal self-organizing method---L\'evy search, threshold recruitment, density-dependent dispersal---\emph{reaches
the shape of the optimal allocation}, the ideal free distribution, balancing exploration and exploitation
through its cascade cycle, with no learner, no trained parameters, and no payoff or reward in its dynamics. It
reaches this allocation as a self-organized attractor and holds it, re-finding it when the resources move; and
under matched embodiment it is the most gradient-faithful of the arms tested and matches the best trained
learner's harvest where resource is clustered (trailing where it is scarce), while learning buys precision and
transfer, not a better allocation \emph{shape}. The reinforcement
learning is at the level of the problem solved and the allocation reached---a result about how little a system
needs to land where a learner is trained to go.

\section*{Statements and Declarations}

\paragraph{Funding.}
No funds, grants, or other support was received for conducting this study.

\paragraph{Competing interests.}
The author has filed a provisional patent application related to the recruitment-switching mechanism described
in this work. The author declares no other competing interests.

\paragraph{Data and code availability.}
Complete simulation code, all configuration files, precomputed sweep and comparison data, figure-regenerating
scripts, and an interactive browser-based demo of the mechanism are released publicly at
\url{https://github.com/galenwilkerson/self-organizing-reinforcement-learning}.

%
\PaperBibliography

\clearpage
\appendix
\makeatletter
\renewcommand\@seccntformat[1]{Appendix~\csname the#1\endcsname\quad}
\makeatother
\section{Complete model specification}
\label{app:modelspec}
We specify the model in full here, so the paper is reproducible without the companion. All constants
are in Table~\ref{tab:params}; the embodied-comparison operating points and the learner setup are in
\appref{app:embodied}. Code and configuration are released (Data and code availability).

\paragraph{Arena and agents.} $N{=}200$ agents live on a square torus $[0,L)^2$, $L{=}800$, with positions
initialised uniformly at random. Time is discrete; each step is one movement plus one recruitment pass. The
embodied comparison runs a horizon $T{=}600$ steps.

\paragraph{Resource field.} There are $C{=}10$ resource clumps with centres drawn uniformly on the torus. Clump
richness is heavy-tailed: $M{=}80$ point sources are assigned to clumps by a Pareto-like weighting---weights
$w_c\propto U_c^{-1/(\beta-1)}$ (with $U_c\sim\mathrm{Unif}(0,1]$, exponent $\beta{=}2.2$) normalised to a
distribution over clumps, then each source assigned by an independent draw---so a clump's richness $Q_c$ (its
number of sources) follows a Pareto-like law over a ${\sim}40$-fold range. Each source is placed at its clump
centre plus an isotropic Gaussian offset of standard deviation $\mathrm{clump\_radius}=\rho_{\mathrm{patch}}L$
(the patchiness $\rho_{\mathrm{patch}}$ is an operating-point parameter; \appref{app:embodied}), wrapped onto
the torus. Every source holds capacity $\kappa{=}40$.

\paragraph{Two interaction ranges.} The model uses two radii. (i) The \emph{proximity graph} joins two agents
within $r=L\sqrt{\rho_r/N}\approx82$ ($\rho_r{=}2.1$); this graph carries recruitment, and an agent's degree
$k$ is its number of neighbours on it. (ii) The \emph{resource-sensing radius} $S=L\sqrt{\rho_s/M}$
sets how close an agent must be to detect a source ($\rho_s$ is the resource coverage, an
operating-point parameter, as in the companion paper; \appref{app:embodied}; $S\approx12.6$ scarce,
$25.3$ clustered). An agent is \emph{on food} when a live source ($\kappa_s>0$) lies
within $S$.

\paragraph{Movement.} Each step an agent takes one of three moves. \textbf{Explore}: a displacement of
heavy-tailed length in a uniformly random direction---the length is drawn from a one-sided $\alpha$-stable law
(Mantegna's algorithm) with exponent $\alpha{=}1.0$ and scale $2.0$. \textbf{Approach}: a \emph{short, local}
step---at the local-search (ARS) scale, not a full L\'evy hop---toward the agent's current \emph{target} (a sensed
source, or, if recruited, the recruitment heading below). \textbf{Stay} (local search): an isotropic Gaussian
jitter of standard deviation $3$. Positions
wrap on the torus. The reward-free swarm's controller chooses explore/approach/stay by the rule below; a
learner replaces that rule with a trained policy over the same three moves (\appref{app:embodied}).

\paragraph{Recruitment cascade (linear threshold, multi-hop).} Recruitment is a synchronous linear-threshold
cascade on the proximity graph, run to a fixed point each step. The \emph{seeds} are the on-food agents. An
inactive agent $i$ \emph{activates} when the active fraction of its neighbours reaches the threshold,
\[
  \frac{a_i}{k_i}\;\ge\;\phi,\qquad \phi=0.18,
\]
where $a_i$ is the number of active neighbours and $k_i$ the degree; rounds repeat until no new agent
activates. A recruited agent (active
but not itself on food) takes the \emph{approach} move---short, local steps toward the resource---along the centroid heading of its \emph{parents}---its
active neighbours that activated in an earlier round---so a discovery is carried outward through the network to
agents that cannot sense it. These recruited steps are short (local-search scale), shorter than an exploring L\'evy
hop, so no extra movement budget is spent to coordinate.

\paragraph{Density-dependent dispersal (blocking).} Blocking is not a separate rule; it is a consequence of the
threshold being \emph{degree-normalized}. As a neighbourhood crowds, each agent gains neighbours faster than it
gains \emph{active} neighbours, so $a_i/k_i$ falls back below $\phi$ and no further agents are recruited to
the crowded region: the same rule that spreads a cascade stifles it once a patch is worked, and the swarm
releases agents back to exploration.

\paragraph{Consumption, congestion, and depletion.} An on-food agent draws from the single nearest live source
it is within $S$ of. The resource a source yields in a step is shared equally among the agents drawing on it:
with $n$ agents on a source of capacity $\kappa_s$, the source yields $\min(\kappa_s,\,c\,n)$ with per-agent
consumption $c{=}1$, so each of the $n$ agents receives $\min(\kappa_s,cn)/n$ and the source's capacity is
decremented by the amount yielded. Per-capita intake therefore falls as a source crowds---the congestion that
makes the optimum an allocation. When a source's capacity reaches $0$ it \emph{respawns} immediately: it is
re-placed at a fresh Gaussian offset around its clump and its capacity is reset to $\kappa$, so the field is
stationary in distribution while individual sources deplete and move.

\paragraph{Branching ratio $\sigma$.} We summarise cascade spread by the branching ratio $\sigma$: the mean
number of secondary activations triggered per active agent over a cascade. $\sigma<1$ is \emph{subcritical}---cascades are finite and die
out; $\sigma\approx1$ is critical. Across the swept density range the mobile swarm holds $\sigma\approx0.067$--$0.086$
(Fig.~\ref{fig:selforg}), i.e.\ globally subcritical.

\paragraph{Per-step update (summary).} (1) sense sources within $S$ and build the proximity graph within $r$;
(2) run the linear-threshold cascade to its fixed point to label on-food/recruited agents and recruitment
headings; (3) choose each agent's move (swarm rule, or learned policy); (4) apply moves with torus wrapping;
(5) resolve shared consumption, decrement capacities, and respawn depleted sources. The configuration carried by
the agents' positions and the proximity graph---stored nowhere else---\emph{is} the swarm's allocation at that
step.

\paragraph{Dimensionless form.} Every raw scale above enters the dynamics only through dimensionless
combinations, so the model is governed by six dimensionless controls (Table~\ref{tab:controls})---the axes of the robustness sweep
(\appref{app:robust}). Three are the swarm's \emph{own} controls, held at task-independent defaults
($\rho_r,\phi,\alpha$); and three are properties of the resource \emph{field}
($\rho_s,\rho_{\mathrm{patch}},\beta$).
\begin{table*}[tp]\centering{\small
\begin{tabular}{@{}llcc@{}}
\hline
control & meaning & operating value & swept range \\
\hline
$\rho_r=N(r/L)^2$ & connectivity (mean degree) & $2.1$ & $[1.0,5.0]$ \\
$\phi$ & activation threshold ($a/k\ge\phi$) & $0.18$ & $[0.05,0.35]$ \\
$\alpha$ & L\'evy exponent & $1.0$ & $[0.6,2.0]$ \\
$\beta$ & Pareto patch-richness exponent & $2.2$ & $[1.5,3.5]$ \\
$\rho_s=M(S/L)^2$ & resource coverage & $0.02$--$0.08$ & $[0.005,0.08]$ \\
$\rho_{\mathrm{patch}}$ & resource patchiness (clump radius$/L$) & $0.0375$ & $[0.02,0.30]$ \\
\hline
\end{tabular}}
\caption{\textbf{The six dimensionless controls} governing the model; \appref{app:robust} sweeps them. Three are the swarm's own controls ($\rho_r,\phi,\alpha$), three are properties of the resource field ($\rho_s,\rho_{\mathrm{patch}},\beta$).}
\label{tab:controls}
\end{table*}

\section{The optimal allocation: ideal free distribution and habitat-game Nash}
\label{app:optimum}
With resource $Q_i$ shared among the $N n_i$ foragers on patch $i$, per-capita intake is $f_i=Q_i/(N n_i)$.
An allocation is stable against unilateral relocation when per-capita intake is equal across all occupied
patches, $f_i=\bar f$ for all occupied $i$. With $f_i=Q_i/(N n_i)$ this gives $n_i\propto Q_i$: the ideal free
distribution~\citep{fretwell1969ideal}. Because each agent's payoff depends only on how many share its patch,
the habitat-selection game is an anonymous congestion game, whose pure Nash equilibrium is exactly this
equal-intake allocation~\citep{rosenthal1973congestion,wardrop1952road}; player-specific congestion games need
not share this structure~\citep{milchtaich1996congestion}. Note that mean per-capita intake
$\langle f\rangle=\sum_i n_i f_i=\sum_i Q_i/N$ is constant on the interior, so the optimum is an equal-intake
\emph{equilibrium} (a proportional-fair, log-utility allocation~\citep{kelly1998rate}), not a total-intake
maximum: the swarm load-balances, it does not increase total intake. We use the ideal free distribution purely
to \emph{characterise} the solution the swarm reaches (slope-one matching, $n_i\propto Q_i$); the swarm does not
compute payoffs or solve the game (\S\ref{sec:how}).

\paragraph{The equilibrium is a potential-game minimiser, and the swarm holds it as an attractor.} The
anonymous congestion game above is a \emph{potential game}~\citep{rosenthal1973congestion,monderer1996potential}:
under equal sharing the unique interior critical point of its exact potential is the equal-intake allocation
$n_i^\ast\propto Q_i$---the ideal free distribution. This fixes the \emph{target} of the dynamics as well-posed
and unique, independently of agent behaviour; it does not imply the swarm reaches it by optimising the potential.
What we establish is instead a \emph{stability} property of the swarm's measured coarse dynamics: the allocation
is reached within ${\sim}200$--$260$ steps and then \emph{held}, without drift, across a $33\times$ density range
(\S\ref{sec:solve}, Fig.~\ref{fig:selforg}), and the mean-field law that best fits the transient
(\appref{app:dynamics}) relaxes \emph{toward} $n_i^\ast\propto Q_i$---so the ideal free distribution is an
attractor of the empirical dynamics that coincides with the potential game's equilibrium.

\section{What the coarse dynamics are: model competition and the role of blocking}
\label{app:dynamics}

\begin{figure*}[tp]\centering
  \includegraphics[width=0.7\linewidth]{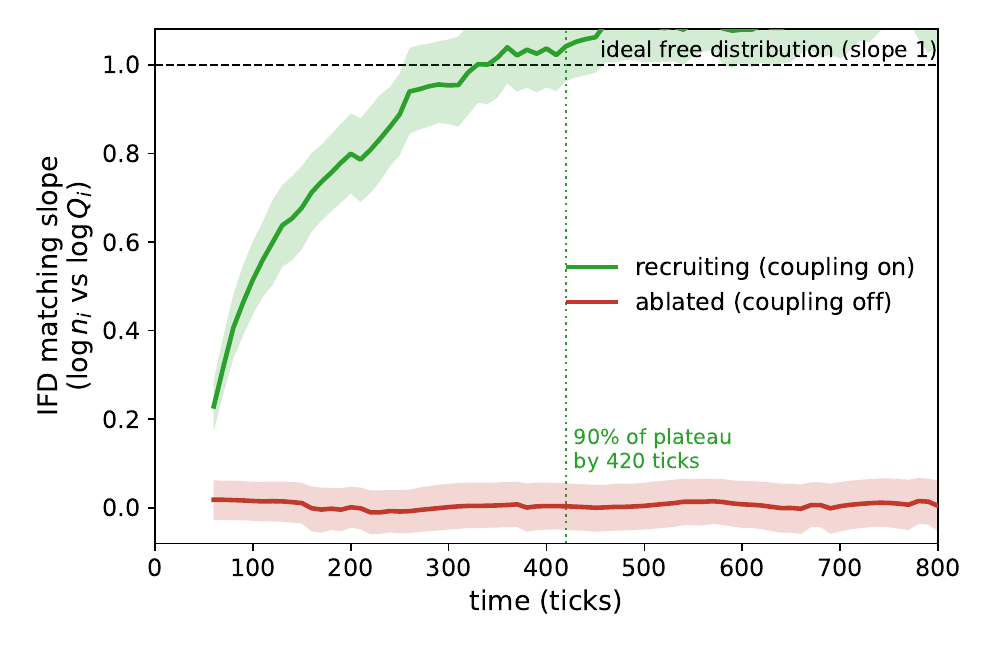}
  \caption{\textbf{The allocation is reached dynamically.} Matching slope (regression of log committed foragers
  per patch on log richness, trailing window; replicas with bootstrap $95\%$ CI) vs.\ time. With the
  recruitment coupling (green) the swarm climbs to its ideal-free plateau (slope slightly above $1$; dashed marks
  the ideal $1$), reaching near-ideal matching by ${\sim}200$--$260$ steps and holding; without it (red) matching
  never develops (slope ${\approx}0.01$).}
  \label{fig:convergence}
\end{figure*}

\paragraph{Model competition.} We ask which mean-field law best describes the measured flow of the allocation
$\dot n_i$. Per-step $\dot n_i$ is dominated by discrete-event noise (all candidate laws sit at held-out
$R^2\approx0.03$ at step resolution), so we fit the \emph{denoised mean flow}: samples binned by
(richness stratum $\times$ occupancy), each model fit to the per-bin mean $\dot n_i$ over the transient
(steps $1$--$150$), with model selection by leave-$16$-replicas-out cross-validation and AIC/BIC. Candidates:
a reward-proportional \emph{replicator} $\dot n_i=r_1\,n_i(f_i-\bar f)$ (which on equal sharing is
algebraically a \emph{linear relaxation} to the fixed point, $r_1'(n_i^*-n_i)$, $n_i^*=Q_i/\sum_jQ_j$);
a saturating \emph{best-response}/logit $\dot n_i=r_2\,(\mathrm{softmax}_\tau(f)_i-n_i)$; a payoff-blind
\emph{logistic/contagion} $\dot n_i=r_3\,n_i(c_0 Q_i-n_i)$; and a constant null. (Here $r_1,r_2,r_3,c_0$ and the
logit inverse-temperature $\tau$ are fitted rate constants, distinct from the model parameters of
\appref{app:modelspec}.) Held-out $R^2$: logit
$0.64$, contagion $0.47$, replicator/linear-relaxation $0.26$, null $0$. Two honest conclusions follow. (i) The
linear replicator/relaxation form is \emph{rejected} as sufficient---the flow has saturating curvature. (ii)
The best-fitting forms are nonlinear \emph{relaxations} toward the $n_i\propto Q_i$ fixed point; because on
equal sharing payoff-rank equals gap-rank, the logit fit does not indicate that the swarm computes payoffs (it
cannot---\S\ref{sec:how}). The flow is a nonlinear relaxation of the allocation to the equilibrium, consistent with
a reward-free mechanism. Two caveats keep this honest. First, under equal sharing the replicator regressor
$n_i(f_i-\bar f)$ equals the signed distance to the fixed point, $(Q_{\rm tot}/N)(n_i^*-n_i)$ (measured
correlation $0.997$), so payoff-rank and gap-rank coincide and \emph{no} regression on this data can separate a
reward-free from a reward-following rule---reward-freeness rests on the rule's construction (\S\ref{sec:mech}),
not on these fits. Second, the payoff-blind logistic/contagion form, though competitive in bulk $R^2$, is
itself rejected by a shape diagnostic: at small occupancy the measured flow grows with the gap rather than
being throttled toward zero as $n_i(c_0 Q_i-n_i)$ would require. So the fits identify ``nonlinear relaxation to
the fixed point'' but neither a specific reward-following law nor a specific payoff-blind one.

\begin{figure*}[tp]\centering
  \includegraphics[width=\linewidth]{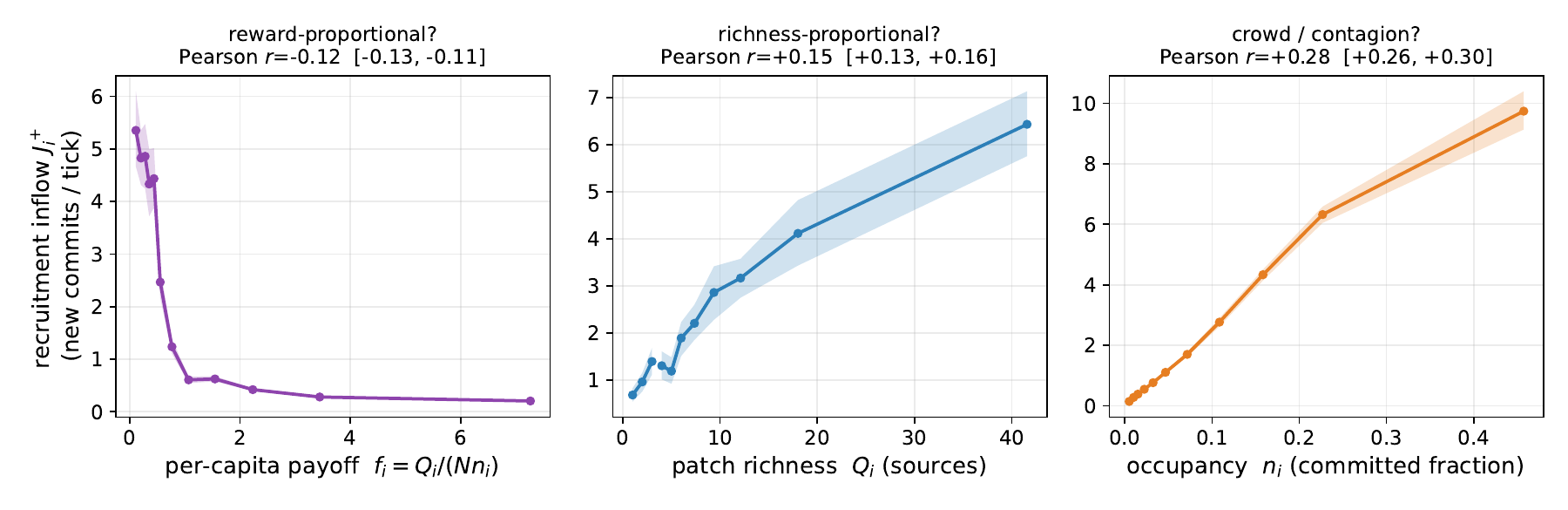}
  \caption{\textbf{Illustrative: recruitment tracks crowd, not payoff} (reward-freeness itself is established by
  the rule's construction, \S\ref{sec:mech}, not by this figure). Per-patch recruitment inflow $J^+_i$ (new
  commits per step) vs.\ per-capita payoff $f_i$, richness $Q_i$, and occupancy $n_i$ (quantile-binned mean,
  bootstrap $95\%$ CI over $64$ replicas). Inflow correlates positively with occupancy ($r=+0.28$)---the crowd
  signal the cascade rule uses---weakly negatively with payoff ($r=-0.12$), and only weakly positively with
  richness ($r=+0.15$, well below occupancy). The correlations are small and, under equal sharing,
  non-identifying; they illustrate rather than prove the reward-free construction.}
  \label{fig:rewardfree}
\end{figure*}

\begin{figure*}[tp]\centering
  \includegraphics[width={\ifarxivbuild 0.72\linewidth\else\linewidth\fi}]{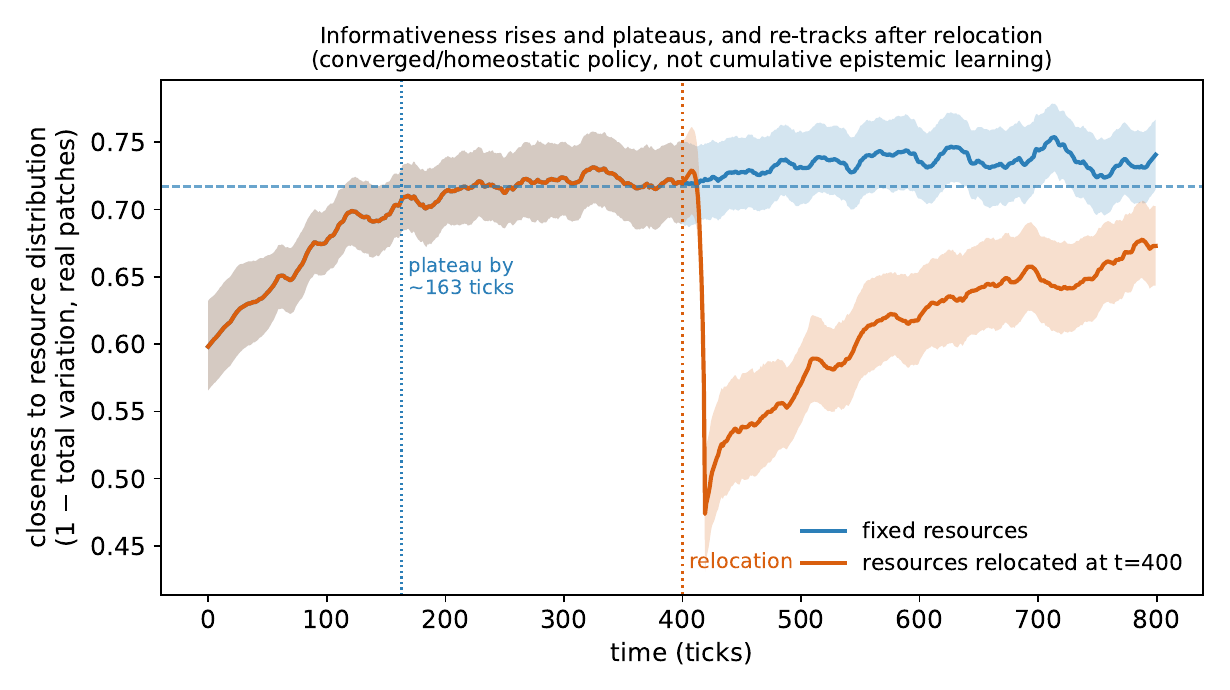}
  \caption{\textbf{The swarm reaches the allocation but does not learn it} (support for \S\ref{sec:how}).
  Allocation informativeness
  (closeness $=1-$ total-variation distance between committed occupancy and the resource distribution; replicas
  with bootstrap $95\%$ CI) vs.\ time. It rises once and \emph{plateaus} by ${\sim}163$ steps; after the
  resources are relocated ($t{=}400$, dashed) it dips and \emph{re-tracks} toward the same plateau---never a
  higher one---no cumulative improvement. A converged policy, not a learner.}
  \label{fig:plateau}
\end{figure*}

\paragraph{A coarse model with the ideal free distribution as a stable fixed point.}
The model competition above identifies the \emph{form} of the flow (a nonlinear relaxation toward
$n_i\propto Q_i$); we now write the minimal coarse model that produces it and show its fixed point is
\emph{linearly stable}, which is the precise sense in which the allocation is an attractor. Let $n_i$ be
the fraction of agents committed to patch $i$ and $u=1-\sum_j n_j$ the exploring pool. The recruitment
cascade seeds patches in proportion to discovery---richer patches, with more sources, are seeded more
often---recruiting from the exploring pool, while degree-normalized blocking releases agents from crowded
patches (\appref{app:modelspec}). The minimal balance of these two terms is
\begin{equation}
  \dot n_i \;=\; a\,Q_i^{\,s}\,u \;-\; b\,n_i^{\,p},
  \qquad u=1-\textstyle\sum_j n_j,
  \label{eq:mf}
\end{equation}
with rate constants $a,b>0$, discovery exponent $s$, and dispersal exponent $p\ge 1$. The inflow is
independent of $n_i$, as the shape diagnostic above requires (at small occupancy the measured flow grows
with the richness gap, not throttled by $n_i$, ruling out an $n_i(c_0Q_i-n_i)$ inflow). The interior fixed
point of Eq.~\eqref{eq:mf} is
\begin{equation}
  n_i^\ast=\Big(\tfrac{a\,u^\ast}{b}\Big)^{1/p} Q_i^{\,s/p},
  \label{eq:mffp}
\end{equation}
so the log--log allocation slope is $s/p$; linear discovery and dispersal ($s=p=1$) give exactly the ideal
free distribution $n_i^\ast\propto Q_i$ (slope one). Linearising Eq.~\eqref{eq:mf} about $n^\ast$ gives the
Jacobian $J=-\big(D+a\,\tilde q\,\mathbf 1^{\!\top}\big)$ with $D=\mathrm{diag}(b\,p\,n_i^{\ast\,p-1})\succ0$
and $\tilde q_i=Q_i^{\,s}>0$. Its eigenvalues solve $1+a\sum_i \tilde q_i/(d_i-\lambda)=0$; the roots
interlace the positive poles $d_i$ and the largest exceeds $\max_i d_i$, so every eigenvalue of $J$ is
\emph{real and negative}. For linear dispersal ($p=1$) this is closed form: the eigenvalues are $-b$ (with
multiplicity $K-1$) and $-(b+a\sum_i Q_i^{\,s})$. The ideal-free fixed point is therefore a
\emph{linearly stable node} of the coarse dynamics~\citep{strogatz2015nonlinear}. Numerically, on the measured ten-patch field the
Jacobian eigenvalues are all real and negative in both resource fields, and integrating Eq.~\eqref{eq:mf}
from random interior initial occupancies converges to $n^\ast$ in every run (code released). This upgrades
the attractor claim from the curve-fit relaxation above to a derived local-stability statement: the
congestion-game equilibrium (\appref{app:optimum}) is a stable fixed point of the swarm's mean-field
dynamics, not merely the target the fits relax toward.

Two limits bound this. The rate constants $a,b$ are not calibrated, so only the \emph{sign} structure (all
eigenvalues negative) is claimed, not the relaxation rate (the ${\sim}200$--$260$-step timescale is the measured
one, \S\ref{sec:solve}). And the slope is \emph{localized}, not predicted from first principles: the product form $s/p$
equals the measured slope (${\approx}1.14$). Because dispersal is at least linear ($p\ge1$) and the measured
slope exceeds one, the discovery exponent satisfies $s\ge1.14$---discovery is super-linear in richness, as
expected when a richer patch simply has more resource sites to stumble onto---but we do not pin the individual
values of $s$ and $p$ beyond this bound. The cutoff-removal ablation below shifts the slope ($1.14\to0.95$) and loosens the
match rather than leaving it fixed, so the self-limiting dispersal also contributes to the precise slope; the
ablation establishes that the order-one matching \emph{survives} removing the cutoff---it is produced by
richness-proportional discovery and recruitment, not created by the cutoff. This is a mean-field,
population-average stability result, not a convergence theorem for the microscopic stochastic reward-free rule
(Discussion, limitation~iv).

\paragraph{The role of density-dependent dispersal (blocking).} We separate blocking's two possible roles:
setting the explore--exploit \emph{balance} versus setting the allocation \emph{shape}. For the balance,
blocking is causal: we show in-paper that the swarm self-tunes to a subcritical operating regime across a
$33\times$ density range with no per-condition tuning (\S\ref{sec:solve}, Fig.~\ref{fig:selforg}), and the
companion paper maps the bounding cascade window in full with phase diagrams over connectivity and activation
threshold (percolation onset below, degree-dependent blocking above)~\citep{partI}. For the allocation shape,
we run a single
intermediate ablation here: removing the crowding cutoff (replacing the degree-normalized rule $a/k\ge\phi$ with
an absolute-count rule, the maximal no-cutoff limit) while leaving recruitment otherwise intact. The matching
slope stays near-ideal and does not collapse ($1.14\to0.95$, no over-aggregation---the slope does not steepen),
while the active fraction rises ($0.71\to0.86$) and the match loosens (total-variation distance to the
ideal-free allocation $0.22\to0.37$). So blocking regulates the explore--exploit balance and sharpens the match,
but does not create the proportional allocation; that shape traces to richness-proportional discovery and
recruitment (richer patches have more resource sites, so are seeded more often). This is a single-operating-point
check, not a sweep---the self-tuning across conditions is Fig.~\ref{fig:selforg} and the full parameter map
is~\citep{partI}---and the absolute-count toggle is an extreme limit, but together they attribute the allocation
shape to the recruitment--movement coupling and discovery, and the explore--exploit balance to self-limiting
dispersal.

\section{Trained multi-agent RL baseline (abstract congestion game)}
\label{app:marl}

\begin{figure*}[tp]\centering
  \includegraphics[width=\linewidth]{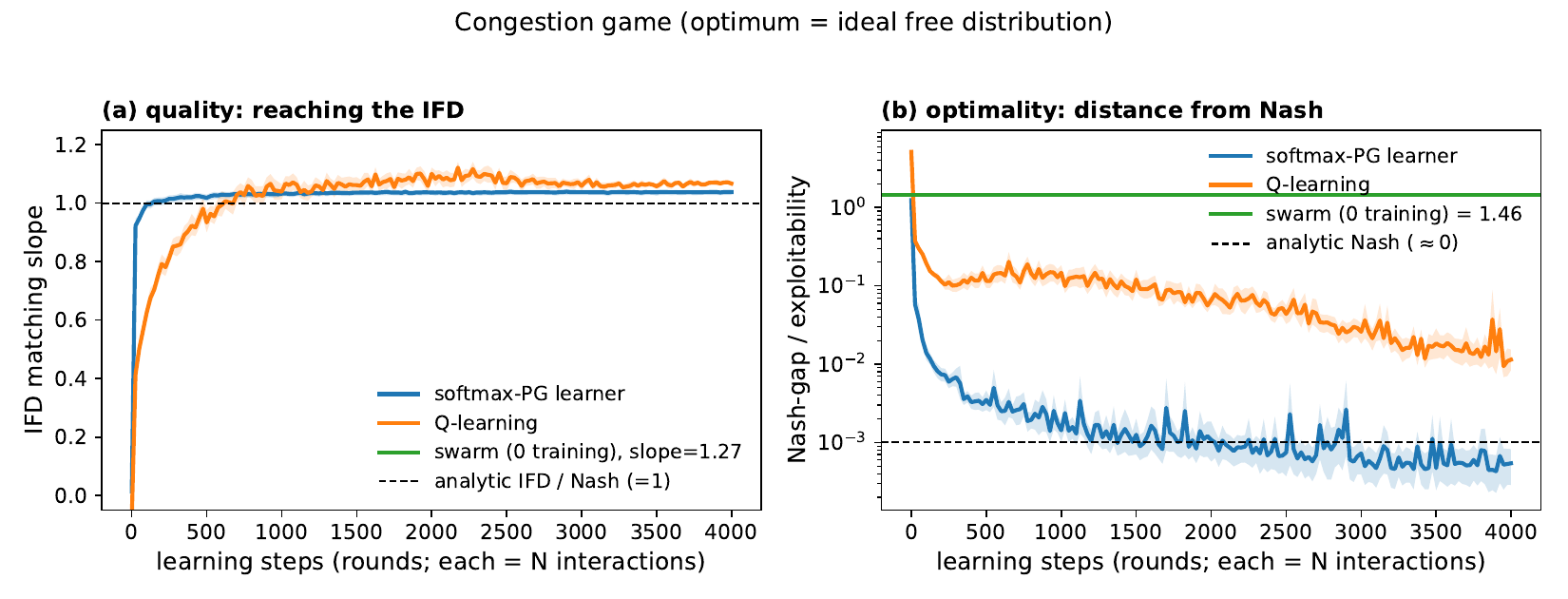}
  \caption{\textbf{Learner-free swarm vs.\ trained multi-agent RL in the abstract congestion game.}
  \textbf{(a)} matching slope and \textbf{(b)} Nash-gap (exploitability) vs.\ learning steps for independent
  softmax policy-gradient and $Q$-learning (bootstrap $95\%$ CI over $64$ replicas), with the zero-training
  swarm and the analytic optimum as references: the swarm reaches the right allocation \emph{shape} at zero
  training (a), but its Nash-gap ($1.46$) stays well above the trained learners' (b)---it is not near-optimal on
  exploitability. The learners play an easier, disembodied problem (no spatial search); units differ (learner
  steps vs.\ swarm physical steps), as marked. The transfer result (congestion vs.\ non-congestion regime) is
  Fig.~\ref{fig:transfer}.}
  \label{fig:marlabstract}
\end{figure*}

On an abstract version of the congestion allocation game ($K$ patches, $N$ agents, per-capita payoff
$f_i=Q_i/(N n_i)$, richness $Q_i$ matched to the swarm's), each agent runs an independent learner with local,
own-payoff information only: a softmax policy-gradient (REINFORCE with a per-agent baseline) or an
$\epsilon$-greedy $Q$-learner. Quality is the matching slope and the distance to the ideal free distribution;
the ``regret'' analogue is the \emph{exploitability} / Nash-gap, since reward-regret is ill-defined here (mean
per-capita intake is constant, \appref{app:optimum}). The swarm, being reward-blind, is run at its physical
operating point and its steady-state allocation compared on the same $(Q_i,N,K)$. The non-congestion regime
replaces the payoff with a fixed per-patch reward $f_i=Q_i$ (optimum $=$ concentrate on $\arg\max_i Q_i$),
where the swarm's fixed mechanism is sub-optimal while the learners adapt (Fig.~\ref{fig:transfer}). Learner
hyperparameters are lightly tuned so the baseline is neither crippled nor over-fit; the swarm has no tunable
parameters.

\section{Embodied learner comparison}
\label{app:embodied}
All arms share one spatial environment: a torus with $N=200$ agents per replica, a local sensing radius
$S$, Pareto-clumped resources that deplete and respawn, and the action set $\{$explore (a L\'evy step),
approach, stay (local search)$\}$. \textbf{Matched, sensing-limited primitives.} The \emph{approach} action
steers an agent only toward a resource it senses \emph{within} $S$ (or, if recruited, along its active
neighbours' heading); no agent can head toward resource beyond its own sensing range. This is identical for
every arm. This matched sensing limit is what makes the comparison a test of \emph{coordination}: it applies
equally to both arms, and recruitment is precisely the swarm's fixed answer to the sensing gap, so the contest
asks whether a learner can learn an equivalent. The result brackets from both sides. (An earlier version let
\emph{approach} target the nearest live resource at \emph{any} range---a clairvoyant primitive, shared by both
arms. Because every agent could then walk straight to unsensed resource, recruitment was moot and the
reward-reading learner \emph{out}-harvested the swarm; that setting tests nothing about coordination, which is
why the sensing-limited numbers supersede it. So with teleportation the learner wins and under matched sensing
they tie where resource is clustered (the learner still wins where it is scarce)---the matched-sensing regime being the one that isolates what recruitment buys.) The \emph{swarm} arm applies the
fixed recruitment$+$blocking rule over this observation/action set, reward-free; each \emph{learner} arm
replaces the rule with a trained policy over the same action set. Only the controller differs---sensing,
movement, and actions are identical---so the comparison isolates fixed versus learned control under matched
embodiment and partial observability, and because both arms use the same movement primitives and step budget
the fairness does not rely on the companion paper's movement-neutrality claim. The swarm's fixed parameters are
moreover \emph{not} tuned to this task: the L\'evy exponent $\alpha{=}1.0$ and the activation threshold
$\phi{=}0.18$ are held at their a-priori values across all conditions and are demonstrably \emph{not}
harvest-optimal (a higher $\phi$ and a heavier-tailed $\alpha$ each raise intake), so parameter selection does
not favour the swarm.

\paragraph{Operating points.} We evaluate on two resource fields: \textsc{scarce} (low resource density, more
dispersed) and \textsc{clustered} (resource-rich, tightly patched---the ecologically typical heavy-tailed case).
The two differ only in environment/sensing parameters (resource coverage $\rho_s$, clump radius, congestion
strength); $\alpha$ and $\phi$ are identical in both.

\paragraph{Learners.} We train two policies by independent PPO (clipped surrogate, generalised advantage
estimation, entropy bonus, a shared-weight actor with a per-agent local critic). \emph{MLP-IPPO} is a
feedforward policy on the egocentric observation. The \emph{communication-capable} policy (GNN) runs $L$ rounds
of learned message passing over the same proximity graph the cascade uses before acting ($L\in\{1,3,5\}$), so it
can in principle propagate a discovery across the network within a step as the cascade does. Both see a rich
$14$-dimensional egocentric observation (on-resource flag; nearest sensed-resource proximity and direction;
local and wide resource density; neighbour-foraging fraction and local degree; nearest on-food neighbour
proximity and direction; own short-memory intake and on-food fraction)---at least as informative as what the
swarm effectively uses, so the learner is advantaged on information, not handicapped. Training uses a dense
shaped reward (per-step per-capita intake plus a small resource-proximity term); \emph{evaluation always uses
the true intake fraction of the oracle ceiling} $\min(N,\sum\text{cap})$ per step, never the shaped reward.
Hyperparameters (shared): learning rate $3\times10^{-4}$, hidden width $64$, rollout $128$, $4$ epochs,
clip $0.2$, $\gamma=0.99$, $\lambda=0.95$, value coefficient $0.5$, entropy $0.01$, gradient-norm clip $0.5$,
Adam.

\paragraph{Shaping/entropy ablation.} The shaped reward's resource-proximity term ($0.05$) and the entropy
bonus ($0.01$) are both mild spreading pressures, so a natural worry is that they---not trained RL---cause the
under-concentration. We retrain the binding comparator (MLP-IPPO) across four cells---full shaping (baseline),
no proximity term, no entropy bonus, and neither---$16$ seeds each, same $150{,}000$-step budget. The matching
slope is essentially unchanged and never approaches the swarm's ${\approx}1$: on the scarce field
$0.76/0.77/0.78/0.76$ and on the clustered field $0.92/0.90/0.89/0.86$
(baseline\,/\,no-shaping\,/\,no-entropy\,/\,neither). Removing the spreading pressures does not raise the slope---on
the clustered field it slightly \emph{lowers} it ($0.92\to0.86$), the opposite of the shaping-artefact
prediction. The learners' under-concentration is therefore intrinsic to what they represent and optimize under
matched sensing, not an artefact of the reward shaping; the swarm's own gradient-unbiasedness (slope ${\approx}1$)
has no reward to shape at all. (The capacity/learning-rate sweep below likewise holds shaping and entropy
fixed.)

\paragraph{Efficient GPU implementation.} The whole environment is a batched GPU (PyTorch) port of the NumPy
environment; every independent unit---$16$ seeds $\times$ $2$ operating points $\times$ learner family---is
folded into a leading ensemble dimension with grouped, per-policy weights, so the full comparison runs as one
high-utilisation GPU job rather than a sequential sweep. The port is validated against the original NumPy
environment: the reward-free multi-hop swarm reaches the same oracle fraction within seed noise and the same
ideal-free slope, and random actions match.

\begin{figure*}[tp]\centering
  \includegraphics[width=\linewidth]{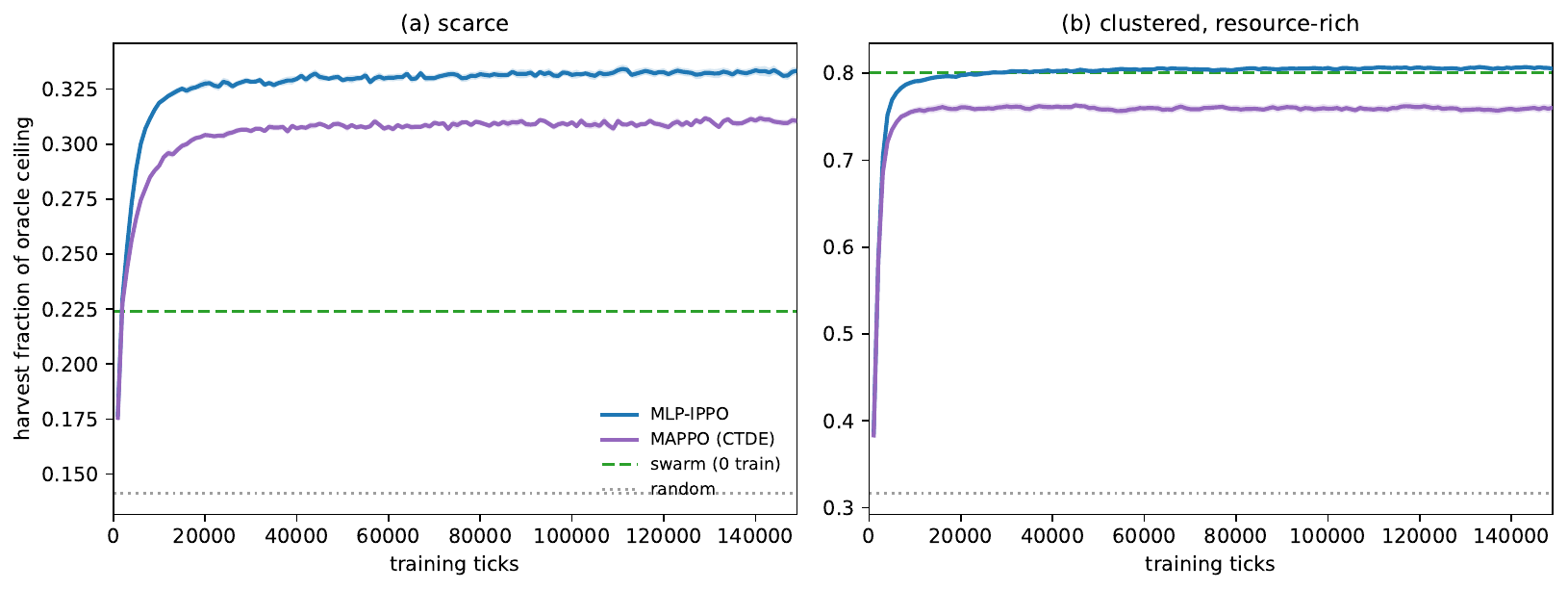}
  \caption{\textbf{The learners are trained to convergence (fairness check).} Harvest fraction of the oracle
  ceiling vs training ticks for the embodied learners (MLP-IPPO and the MAPPO centralized-critic steelman), with
  the learner-free swarm and the random baseline as horizontal references, in the \textbf{(a)} scarce and
  \textbf{(b)} clustered fields ($16$ seeds; bands are $\pm$SEM over seeds). Both learners plateau well before the
  $150{,}000$-step budget---MLP-IPPO's relative gain over the final quarter is ${<}0.3\%$ in both fields, slope
  ${\approx}2\times10^{-5}$ per $1000$ ticks over the last half---so the learner had every opportunity to
  converge before being compared to the zero-training swarm.}
  \label{fig:learncurves}
\end{figure*}

\paragraph{Protocol and result.} Each learner is trained for $150{,}000$ environment steps---$10\times$ the
earlier budget, long enough that the binding comparator (MLP-IPPO) is plateaued over the last half of training
(relative gain ${<}0.3\%$ over the final quarter in both fields; Fig.~\ref{fig:learncurves})---and scored on a
trailing evaluation window; we run $16$ seeds per policy (the seed is the unit of uncertainty, since the
shared-weight policy couples replicas) and compare against multi-seed reward-free swarm and random-action
references on the same fields. We evaluate a third learner family as a steelman: MAPPO, a
centralized-training--decentralized-execution learner whose actor is the identical sensing-limited IPPO actor but
trained against a centralized critic that sees privileged global state (so the execution-time comparison stays
fair while the training signal is the strongest a fair actor can be given). Fractions of the oracle ceiling
(mean over $16$ seeds; per-seed $95\%$ CIs $\lesssim0.01$ except the higher-$L$ GNN, $\lesssim0.03$) are collected in Table~\ref{tab:eharvest}.

\begin{table*}[tp]\centering\resizebox{\linewidth}{!}{{\small
\begin{tabular}{@{}lcccccc@{}}
\hline
 & random & swarm ($0$ train) & MLP-IPPO & MAPPO (CTDE) & GNN ($L{=}1/3/5$) & GRU \\
\hline
\textsc{scarce}    & $0.14$ & $0.22$ (slope $0.99$) & $0.33$ (slope $0.76$) & $0.31$ (slope $0.69$) & $0.31/0.29/0.26$ & $0.31$ (slope $0.78$) \\
\textsc{clustered} & $0.32$ & $0.80$ (slope $1.02$) & $0.80$ (slope $0.91$) & $0.76$ (slope $0.83$) & $0.78/0.70/0.71$ & $0.75$ (slope $0.97$) \\
\hline
\end{tabular}}}
\caption{\textbf{Embodied harvest, fraction of the oracle ceiling} (mean over $16$ seeds), with ideal-free matching slopes: the reward-free swarm at zero training versus trained MLP-IPPO, MAPPO (CTDE), GNN ($L$ message-passing rounds), and GRU learners, on scarce and clustered fields.}
\label{tab:eharvest}
\end{table*}

(The $L{=}1$ GNN harvest is the matched-environment rerun---$R{=}24$ parallel environments, $24$ seeds, per-seed sd $0.01$ in both
fields; the $L{=}3,5$ harvest points and all GNN allocation-slope and exploitability figures are from the
original $R{=}10$ campaign. This rerun reproduces every other arm's numbers to the stated precision.)

In the clustered field the reward-free swarm's harvest is statistically \emph{equivalent} to the best trained
learner's ($0.80$ vs MLP-IPPO's $0.80$; a two-one-sided-test rejects a difference beyond $\pm0.02$ at $p<0.001$,
$90\%$ CI of the difference $[-0.004,+0.008]$, $16$ seeds) and is not out-harvested by the MAPPO steelman
($0.76$); in the scarce field the learner harvests more ($0.33$ vs $0.22$, $p<0.001$, $16$ seeds). The
allocation result is a bias--variance split. On the allocation \emph{shape} (ideal-free matching slope) the swarm
is the most faithful arm in both fields (slope ${\approx}1$ vs the learners' $0.69$--$0.97$): it is approximately unbiased
(slope ${\approx}1$) with respect to the resource gradient while the learners under-concentrate. On allocation \emph{precision}
(exploitability $=$ occupancy-weighted CV of per-capita intake over patches with $n_i\ge2$) the ranking reverses (Table~\ref{tab:eprec}).

\begin{table*}[tp]\centering{\small
\begin{tabular}{@{}lccccc@{}}
\hline
exploitability (per-capita CV, $n_i\ge2$) & random & swarm & MLP-IPPO & GNN ($L{=}1$) & GRU \\
\hline
\textsc{scarce}    & $0.63$ & $0.42$ & $0.26$ & $0.26$ & $0.25$ \\
\textsc{clustered} & $0.90$ & $0.55$ & $0.36$ & $0.48$ & $0.39$ \\
\hline
\end{tabular}}
\caption{\textbf{Allocation precision (exploitability)}: occupancy-weighted CV of per-capita intake over patches with $n_i\ge2$ (lower is tighter), for the same arms and fields as Table~\ref{tab:eharvest}.}
\label{tab:eprec}
\end{table*}

\paragraph{A recurrent learner (learnable memory).} To test whether \emph{learnable} memory---beyond the
hand-crafted EMA features already in the observation---changes the picture, we add a GRU-IPPO arm: a per-agent
gated-recurrent policy whose hidden state is carried across environment ticks, trained by the same PPO with
truncated backpropagation-through-time over each rollout, on the identical sensing-limited observation/action
set and the same $150{,}000$-step budget ($16$ seeds; it plateaus well before the budget). It falls in the same
pattern as the feedforward learners: clustered harvest $0.75$ (below the swarm's $0.80$ and the feedforward
IPPO's $0.80$) and scarce harvest $0.31$ (above the swarm's $0.22$, like the other learners); it
under-concentrates (slope $0.78$ scarce, $0.97$ clustered, versus the swarm's $0.99$/$1.02$) and equalizes
per-capita intake more tightly (exploitability $0.25$/$0.39$ versus the swarm's $0.42$/$0.55$). Learnable
recurrent memory does not overturn the comparison at this budget (one architecture and training budget, not the
full space of either).

The trained learner equalizes per-capita intake more tightly than the swarm, though the swarm remains well below
random; learning buys precision, not a better gradient match. We report the $n_i\ge2$ exploitability because it is
stable across occupancy thresholds ($n_i\ge0.5/2/4$) and under a Gini measure; the looser $n_i\ge0.5$ floor
inflates the CV on marginally-occupied patches and spuriously lifts the swarm's clustered-field value above
random. The communication-capable GNN does not beat the swarm and underperforms even the feedforward MLP-IPPO,
and the MAPPO centralized critic does not help---so neither communication nor joint credit assignment overturns
the comparison at this budget. Additional message-passing rounds ($L{=}3,5$) are no better (run only at the
original, smaller environment count, so reported as a bound). Re-running the single-round GNN at a matched count
($24$ environments, $24$ seeds) cut its per-seed standard deviation from $0.04$ to $0.01$ and raised its
clustered harvest from $0.755$ to $0.78$---still below the swarm and the feedforward IPPO---so the deficit is not
gradient-variance from too few environments. As correctness checks, the learners rise well above random and the
same PPO climbs the ideal-free matching slope on the easy abstract congestion game (\appref{app:marl}),
confirming the optimiser is sound.

\paragraph{The baseline is not under-powered.} To confirm that the learner's ceiling (${\approx}0.80$ clustered,
${\approx}0.33$ scarce) is not an artefact of an under-capacity or under-tuned network, we retrained the
feedforward MLP-IPPO learner across hidden width ($64\to128\to256$, up to $4\times$ capacity) and learning rate
($3\times10^{-4}\to10^{-3}$), $8$ seeds each. Capacity and tuning move the ceiling by less than $0.01$: the
clustered-field fraction stays $0.80$--$0.81$ (width-$256$ reaches $0.808$) and the scarce-field $0.329$--$0.332$.
The reward-free swarm's clustered-field equivalence ($0.801$) therefore holds against the strongest learner we
trained, within the $\pm0.02$ margin---not merely the default one.

A note on earlier versions: a deliberately simple online REINFORCE controller once plateaued near random
and was mistaken for evidence that an embodied learner cannot match the swarm; a later, properly-trained
comparison instead \emph{over}-credited the learner because it ran under the clairvoyant any-range approach
primitive. The sensing-limited comparison here supersedes both---the swarm matches the best learner's harvest
where resource is clustered (statistically equivalent) and is the most gradient-faithful of the
arms tested in both fields, the learner is more precise on per-capita intake, and neither learned communication
nor a centralized critic changes that.

\section{Robustness of the allocation across the dimensionless controls}
\label{app:robust}
To check that the near-ideal allocation is not an artefact of the single operating point, we ran a
low-discrepancy (Sobol) sweep of $1024$ points over the six dimensionless controls of the proper multi-hop
model---connectivity $\rho_r$, activation threshold $\phi$, L\'evy exponent $\alpha$, patch-richness heterogeneity $\beta$, resource
coverage $\rho_s$, and resource patchiness---each point evaluated over $B{=}64$ independent
replica environments, scoring the ideal-free matching slope and the fraction of the oracle ceiling. Because the box is
wide, most random points fall in poor regimes; the question is whether the operating point sits on a broad,
forgiving plateau in the controls the swarm itself sets, and what the allocation actually depends on.

\begin{figure*}[tp]\centering
  \includegraphics[width=\linewidth]{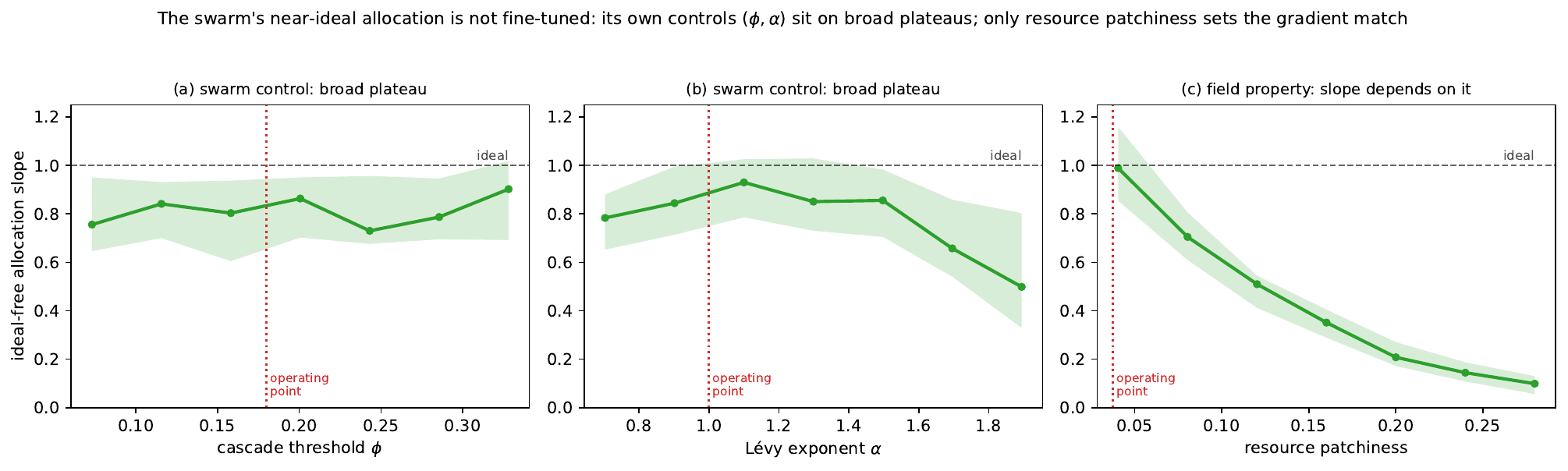}
  \caption{\textbf{The near-ideal allocation is not fine-tuned.} Ideal-free matching slope (binned median, IQR
  band) against three controls, from the $1024$-point Sobol sweep; the dashed line is the ideal slope $1$ and the
  dotted line marks the operating point. \textbf{(a)} The activation threshold $\phi$ and \textbf{(b)} the L\'evy
  exponent $\alpha$ are the swarm's \emph{own} controls: both sit on broad plateaus---the slope is near-ideal
  across the whole swept range of $\phi$ and over a wide band of $\alpha$, with $\alpha{=}1$ (inverse-square
  L\'evy) at the peak---so neither is tuned. \textbf{(c)} Resource \emph{patchiness} is a property of the
  \emph{field}, not a swarm parameter, and it is the one axis the gradient match depends on: the slope degrades as
  clumps spread out, and the operating point sits in the tight-clump region where matching is near-ideal.}
  \label{fig:robust}
\end{figure*}

Two findings follow (Fig.~\ref{fig:robust}). First, the swarm's own controls are \emph{not} fine-tuned: the
matching slope is near-ideal across the entire swept range of the activation threshold $\phi$ ($0.05$--$0.35$) and
over a wide band of the L\'evy exponent $\alpha$, with the inverse-square value $\alpha{=}1$ sitting at the peak
rather than on a cliff. This is the quantitative form of the ``broad plateau'' the main text relies on when it
says $\alpha$ and $\phi$ are held at a-priori values and are not harvest-optimal (\S\ref{sec:solve}). Second, the
one control the gradient match genuinely depends on is the resource \emph{patchiness}---a property of the
environment, not of the swarm: when clumps are tightly concentrated (as in the operating fields) matching is
near-ideal, and it degrades as the field spreads out. This locates the scope of the result honestly: the swarm
reaches the ideal-free allocation when resource is clustered, the common ecological case the paper
targets, and not when it is diffuse. This sweep varies the swarm's \emph{own} allocation slope across the
patchiness axis; the head-to-head swarm-vs-learner \emph{harvest} contest is swept alongside it over the same
${\sim}5$ patchiness points (\S\ref{sec:compare}), tracing where the swarm ties the trained learner and where the
learner's scarce-field edge opens.

\begin{figure*}[tp]\centering
  \includegraphics[width=\linewidth]{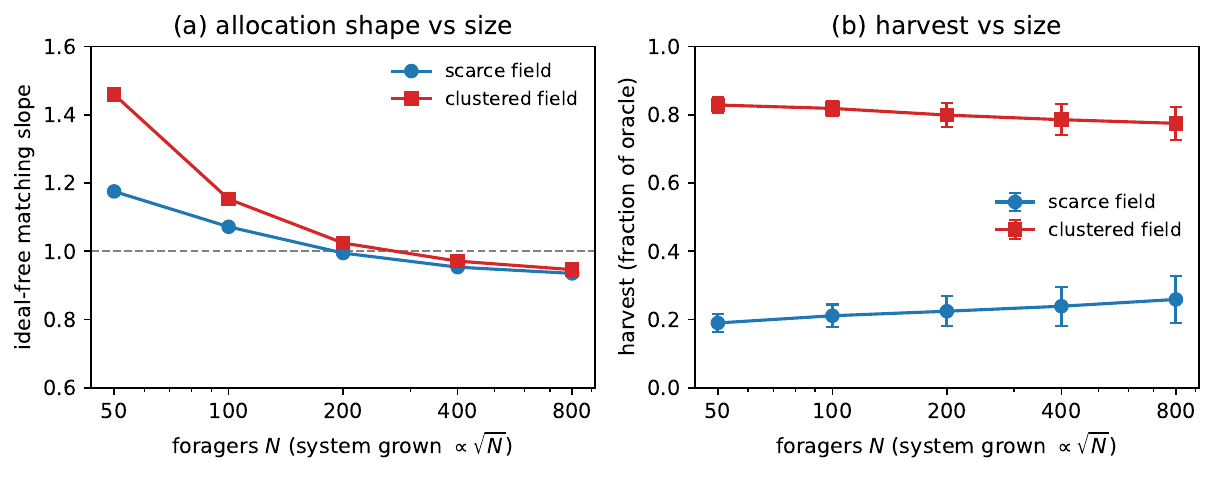}
  \caption{\textbf{The ideal-free matching is robust to system size.} The system is grown with all dimensionless
  controls held fixed ($L\propto\sqrt{N}$, $K,C\propto N$, so $\rho_r$, the interaction radius $r$, every length
  ratio, and every density are constant), over a $16\times$ range of forager count $N$. \textbf{(a)} The matching
  slope stays near one and converges to ${\approx}0.94$ in both the scarce and clustered fields (dashed: slope
  $=1$); the small-$N$ over-concentration damps with size. \textbf{(b)} The harvest fraction (of the oracle
  ceiling) is flat across $N$. The $N{=}200$ point used throughout the paper is representative, not special.}
  \label{fig:nscaling}
\end{figure*}

\paragraph{Robustness to system size.} The main results are at a single forager count ($N{=}200$). To check the
allocation is not an artefact of that scale, we grow the \emph{system} while holding every dimensionless control
fixed: the arena area scales with $N$ ($L\propto\sqrt{N}$) and the resource counts with it ($K\propto N$
sources, $C\propto N$ clumps), keeping $\rho_r$, the interaction radius $r$, all length ratios, and all field
densities invariant---so only absolute size changes. (A fixed-arena $N$-sweep would instead shrink $r$ and drift
the other ratios, measuring those rather than size.) Over a $16\times$ range ($N=50$ to $800$) the ideal-free
matching slope stays close to one and \emph{converges} to ${\approx}0.94$ as $N$ grows---within $[0.94,1.18]$
scarce and $[0.95,1.46]$ clustered---while the harvest fraction is flat (Fig.~\ref{fig:nscaling}). The mild
over-concentration at the smallest $N$ (slope ${>}1$) is a finite-size effect that damps with size; the
$N{=}200$ values used throughout are representative, and on this axis the result settles rather than merely
bounds the size dependence. (Scaling across \emph{learner} architectures and agent counts is not swept, so the
swarm-vs-RL comparison characterises the patchiness axis rather than the full learner/agent-count space.)

\end{document}